\documentclass[letterpaper,table]{article} 
\usepackage{aaai2026}  
\usepackage{times}  
\usepackage{helvet}  
\usepackage{courier}  
\usepackage[hyphens]{url}  
\usepackage{graphicx} 
\usepackage{natbib}  
\usepackage{caption} 
\usepackage{algorithm}
\usepackage{algorithmic}

\usepackage{amsfonts}
\usepackage{amsmath}
\usepackage{amsthm}
\usepackage{booktabs}
\usepackage{algorithm}
\usepackage{algorithmic}

\usepackage{multirow}

\usepackage{tikz}
\usetikzlibrary{shapes.arrows,shapes.geometric, arrows, shadows,positioning}

\usepackage{pgfplots}

\usepackage{paralist}

\usepackage{newfloat}
\usepackage{listings}
\usepackage{listings-rust}
\usepackage{subcaption}

\DeclareCaptionStyle{ruled}{labelfont=normalfont,labelsep=colon,strut=off} 
\floatstyle{ruled}
\newfloat{listing}{tb}{lst}{}
\floatname{listing}{Listing}
\def\defemb#1#2{\expandafter\def\csname #1\endcsname
                              {\relax\ifmmode #2\else\hbox{$#2$}\fi}}
\defemb{cA}{{\cal A}}
\defemb{cB}{{\cal B}}
\defemb{cC}{{\cal C}}
\defemb{cc}{{\cal c}}
\defemb{cD}{{\cal D}}
\defemb{cE}{{\cal E}}
\defemb{cF}{{\cal F}}
\defemb{cv}{{\cal v}}
\defemb{cx}{{\cal x}}

\defemb{cG}{{\cal G}}
\defemb{GG}{{\cal G}}

\defemb{cI}{{\cal I}}
\defemb{cK}{{\cal K}}
\defemb{cP}{{\cal P}}
\defemb{cL}{{\cal L}}
\defemb{cN}{{\cal N}}
\defemb{cM}{{\cal M}}
\defemb{cR}{{\cal R}}

\defemb{cS}{{\cal S}}
\defemb{SS}{{\cal S}}

\defemb{cT}{{\cal T}}
\defemb{cV}{{\cal V}}
\defemb{cU}{{\cal U}}
\defemb{cO}{{\cal O}}
\defemb{cH}{{\cal H}}
\defemb{cW}{{\cal W}}
\defemb{cX}{{\cal X}}
\defemb{cY}{{\cal Y}}
\defemb{cZ}{{\cal Z}}

\defemb{sP}{{\sf P}}
\defemb{sG}{{\sf G}}
\defemb{sU}{{\sf U}}
\defemb{sE}{{\sf E}}
\defemb{sI}{{\sf I}}
\defemb{su}{{\sf u}}
\defemb{se}{{\sf e}}
\defemb{ss}{{\sf s}}
\defemb{st}{{\sf t}}
\defemb{sp}{{\sf p}}
\defemb{si}{{\sf i}}
\defemb{sw}{{\sf w}}
\defemb{sA}{{\sf A}}
\defemb{sB}{{\sf B}}
\defemb{sa}{{\sf a}}
\defemb{scc}{{\sf c}}
\defemb{sd}{{\sf d}}

\title{Successor-Generator Planning with LLM-generated Heuristics}
\author {
    Alexander Tuisov\textsuperscript{\rm 1},
    Yonatan Vernik\textsuperscript{\rm 2},
    Alexander Shleyfman\textsuperscript{\rm 2}
}
\affiliations {
    \textsuperscript{\rm 1}Independent Researcher\\
    \textsuperscript{\rm 2}Computer Science Department, Bar-Ilan University\\
   queldelan@gmail.com, yonatanw55@gmail.com, alexander.shleyfman@biu.ac.il
}

\pgfplotsset{compat=1.18}

\usepackage{bibentry}

\begin{document}

\maketitle

\begin{abstract}
Heuristics are a central component of deterministic planning, particularly in domain-independent settings where general applicability is prioritized over task-specific tuning. This work revisits that paradigm in light of recent advances in large language models (LLMs), which enable the automatic synthesis of heuristics directly from problem definitions -- bypassing the need for handcrafted domain knowledge. We present a method that employs LLMs to generate problem-specific heuristic functions from planning tasks specified through successor generators, goal tests, and initial states written in a general-purpose programming language. These heuristics are compiled and integrated into standard heuristic search algorithms, such as greedy best-first search. Our approach achieves competitive, and in many cases state-of-the-art, performance across a broad range of established planning benchmarks. Moreover, it enables the solution of problems that are difficult to express in traditional formalisms, including those with complex numeric constraints or custom transition dynamics. 
We provide an extensive empirical evaluation that characterizes the strengths and limitations of the approach across diverse planning settings, demonstrating its effectiveness.
\end{abstract}

\section{Introduction}
Deterministic AI planning involves finding a sequence of actions that moves an agent from an initial state to a goal state, given a formal description of the environment's dynamics~\cite{ghallab-et-al:2004}. Because state spaces are usually exponential or even infinite, effective search strategies require heuristics -- functions that estimate the distance to the goal and help prioritize promising states~\cite{pearl:1984,bonet2001planning}. Heuristic search has therefore become a central paradigm in classical planning, underpinning many of the most successful planners developed over the past decades~\cite{hoffman:jair-2003,helmert:jair-2006,scala-et-al:ijcai-2016,aldinger-nebel:socs-2017}. 


Classical planning has traditionally focused on domain-independent heuristics, which offer broad applicability and avoid the need for task-specific design~\citep{bonet2001planning}. However, the performance of such heuristics can vary with the representation language and the structural characteristics of the domain. In more expressive settings, general heuristics may encounter challenges and need adaptation to remain effective across different versions of \textsc{pddl}~\cite{mcdermott-et-al:tr-1998,fox-long:jair-2003,edelkamp-hoffmann:tr-2004} (e.g., cf. classical vs. numeric settings~\citep{hoffman:aimag-2001,hoffman:jair-2003,helmert-domshlak:icaps-2009,kuroiwa-et-al:jair-2022}). For further discussion of current limitations within \textsc{pddl} see \citet{edelkamp:icapsws-2003,rintanen:aaai-20215}.



In this work, we propose a method for \emph{automatically} generating problem-specific heuristic functions from formal problem definitions using LLMs. The generated heuristics are then used by a sound search algorithm, such as greedy best-first search (GBFS). This setup preserves the systematic and provably correct behavior of classical search methods, while making heuristic construction fully automatic and eliminating the need for expert-designed heuristics. 

We represent each planning task using three components, implemented in a general-purpose programming language: a successor generator, a goal test, and an initial state~\cite{russell-norvig:1995,guan-etal:nips-2023,oswald-etal:icaps-2024}. We refer to this representation as an \emph{Explicit Successor Generator} (ESG). These ESG components, together with a prompt, are provided as input to the LLM, which returns a heuristic function (expressed in the same programming language) tailored to the structure of the given problem. The heuristic is then compiled and used to guide search without any further interaction with the LLM.

This approach offers several benefits. It removes the need for repeated LLM calls during planning---a source of significant inefficiency in prior work~\cite{tos:nips-2024,valmeekam-etal:corr-2022,kambhampati-etal:icml-2024}. It also provides transparent search components: both the LLM-generated heuristic and the task specification are expressed in a general-purpose programming language (Rust in this work), allowing inspection, debugging, and testing. In addition, it accommodates planning tasks that involve constructs difficult to model in traditional frameworks, including recursive goals, intricate numeric conditions, the creation of new variables, or custom transition logic.

We show that LLM-generated heuristic can strengthen general-purpose ESG planning. Our evaluation includes a set of standard numerical planning benchmarks represented in the ESG form (using Rust), as well as new benchmarks that are challenging to encode in \textsc{pddl}. Across many domains, the resulting heuristics achieve strong performance, matching the level of state-of-the-art numeric planners in numerous cases. Furthermore, the method is able to solve planning tasks that currently lack formal encodings in existing planning languages.
%




\section{Related Work}

This paper revisits AI planning by heuristic search, by replacing domain-independent heuristics with LLM-generated domain-specific ones. Below, we review related work on LLMs and planning. For a more comprehensive overview of LLMs as planning modelers, see the surveys by \cite{tantakoun-etal:corr-2025} and by \cite{aghzal2025surveylargelanguagemodels}.

\paragraph{Planning with LLMs} \cite{valmeekam-etal:corr-2022} presents an early evaluation of LLM-based planning, showing that even on simple benchmarks, LLMs are not consistently reliable planners. While they cannot yet generate plans efficiently, our work shows they are effective within a broader planning framework, for example by generating heuristics and search components rather than planning directly.

A closely related approach is presented by \cite{tos:nips-2024}, who argue that existing LLM-based planning methods are neither sound nor complete and incur high computational costs due to repeated LLM calls. They propose an alternative, \emph{Thought of Search (ToS)}, in which LLMs are used to generate symbolic search components---successor functions and goal tests. These can then be executed efficiently without further model queries. Although their work focuses on small, finite search problems such as the \emph{24 game} and \emph{Mini Crosswords}, they provide compelling evidence that such an approach can drastically reduce computational cost and improve correctness (solved by BFS). The work shows that with respect to planning their approach overperforms the planning directly with LLM approaches such as: Chain-of-Thought~\citep{wei2022chain}, ReAct~\citep{react:iclr-2023}, ReWOO~\citep{rewoo:corr-2023}, Reasoning via Planning~\citep{rap:emnkp-2023}, Tree of Thoughts~\citep{tot:nips-2023}, Graph of Thoughts~\citep{got:aaai-2024}, Reflexion~\citep{reflection:nips-2023}, Algorithm of Thought~\citep{aot:icml-2024}. The ToS approach runs a breadth-first search (BFS) using components---successor functions and goal tests generated by LLM. Our approach extends this intuition by employing heuristic search methods to address problems of significantly greater complexity than those solvable through naive BFS---specifically in AI planning, where transition systems often become infinite due to the presence of unbounded numeric fluents. Thus, generating effective heuristics is crucial, underscoring that the symbolic utilization of LLMs can be successfully applied beyond toy domains.

A recent study in a similar vein to our work is presented in \cite{correa2025classical}. While their approach targets planning tasks specified in \textsc{pddl}, ours departs from this framework by operating directly on successor generators and goal-check functions. This shift affords greater representational flexibility, enabling the modeling of tasks that \textsc{pddl} cannot express effectively. However, abandoning \textsc{pddl}’s structured representation also limits our ability to exploit domain structure for heuristic derivation.

\paragraph{Heuristic Generation Outside of Planning}
LLMs have also been explored for heuristic generation beyond automated planning, notably in domains such as online bin packing and TSP \cite{liu2024evolution}, as well as other combinatorial optimization problems \cite{ye2024reevo}. These approaches typically combine LLMs with evolutionary algorithms to evolve heuristics that guide the optimization process. However, due to fundamental differences in problem structure, these methods are not directly applicable to planning tasks without substantial adaptation.

\paragraph{Problem Generation} Recent work has explored using LLMs to generate domains and problem descriptions using {\sc pddl}2.1 \citep{fox-long:jair-2003}, among such works are \cite{guan-etal:nips-2023,liu-etal:corr-2023,oswald-etal:icaps-2024}. For instance, \cite{silver-etal:aaai-2024} demonstrated that GPT-4 can output plans from parsed {\sc pddl}2.1 inputs using simple, non-search-based strategies. \cite{guan-etal:nips-2023} proposed an approach in which {\sc pddl}2.1 actions are generated one at a time via LLM queries, refined using human feedback. A related concept appears in \cite{zhou-etal:iclr-2024}, who use LLMs to generate, execute, and refine Python code for solving mathematical reasoning problems. Their findings support the idea that LLMs can produce verifiable code with minimal manual intervention. To reduce human involvement even further, we propose generating Rust code instead of Python. Since Rust is a compiled language with strong static typing, much of the burden of code verification can be offloaded to the compiler.\\
\noindent\textbf{Code Generation} Our approach relies on LLMs not to solve problems directly, but to generate search guidance in the form of state evaluation heuristic code. \cite{zhou-etal:iclr-2024} pursued a similar strategy in mathematical domains, showing that LLMs can generate and iteratively improve executable code. Our work builds on this insight, but targets search guidance functions within a planning framework. The viability of this approach is supported by recent advances in program synthesis and LLM-based code generation \citep{madaan-etal:nips-2023,zhang-etal:acl-2023,chen-etal:iclr-2024,muennighoff-etal:iclr-2024,zhong-etal:aclf-2024}, which show that language models are increasingly capable of producing correct, type-safe, and semantically meaningful programs.

\section{Deterministic AI Planning}

AI planning refers to the problem of computing a sequence of actions that transforms a known initial state into one that satisfies a specified goal condition. Unlike reinforcement learning, which learns behavior through trial-and-error interaction with an environment, AI planning assumes full knowledge of the state transition dynamics and the goal. The planner searches the combinatorial space of states reachable via actions to construct a valid plan.

Formally, a deterministic planning problem can be defined as a tuple $\Pi = \langle V, A, T, s_0, G \rangle$, where:  $V$ is a set of variables with either finite or numeric domains, $A$ is a set of symbolic actions, $T$ is a deterministic transition function, $s_0 \in S$ is the initial state, and $G$ is the goal description. A state $s \in S$ is a full assignment over the variables in $V$. Since the state space $S$ is typically exponential in $|V|$ (or even infinite in the numeric case), it is not explicitly represented in the input. The goal description $G$ is usually represented as a set of conditions. A state $s_\ast \in S$ is considered a goal state if it satisfies $G$, i.e., for each $\varepsilon \in G$ the state $s_\ast$ satisfies $\varepsilon$. The transition function $T$ maps a state-action pair to the resulting state, i.e., $T(s, a) = s'$. A solution (or \emph{plan}) for $\Pi$ is a sequence of actions $\langle a_1, \ldots, a_n \rangle$ such that applying them in order leads from $s_0$ to a goal state $s_n$.

Planning problems can be classified as either \emph{satisficing}, where the goal is to find any valid plan, or \emph{optimal}, where the goal is to find a cost-minimizing plan. In this work, we focus on satisficing planning, which prioritizes informative heuristics over theoretically sound ones (e.g., admissibility).

Solutions to planning problems are typically found via heuristic search algorithms such as GBFS or A*~\citep{hart-et-al:ieee-ssc-1968, pearl:1984}. These algorithms are guided by heuristic functions that estimate the cost or distance from a given state to the goal~\citep{pearl:1984}. A heuristic function $h: S \to \mathbb{R}$ provides such estimates, often derived using domain-independent approximations, such as relaxations of the planning problem. Traditional heuristics are based on symbolic representations and formal action models (e.g., in \textsc{pddl2.1})~\citep{fox-long:jair-2003}, but recent work also explores learned heuristics~\citep{oswald-etal:icaps-2024, chen-etal:aaai-2024}. Informative heuristics greatly reduce the number of explored states, making them essential for practical planning. In this work, we use an LLM to generate a heuristic function based on the formal problem description written in a general-purpose programming language (Rust). This heuristic is integrated into a GBFS planner to obtain satisficing solutions.

\begin{figure*}[t!]
\baselineskip=15pt
\centering
\tikzstyle{startstop} = [rectangle, rounded corners, minimum width=2.5cm, minimum height=1cm,text centered, draw=black, fill=red!30, drop shadow]
\tikzstyle{io} = [rounded corners, text width=18mm, minimum height=13mm, align=center, text centered, draw=black, fill=blue!30, drop shadow]
\tikzstyle{process} = [rounded corners, text width=18mm, minimum height=13mm, text centered, draw=black, fill=orange!30, drop shadow]
\tikzstyle{manual} = [rounded corners, text width=18mm, minimum height=13mm, text centered, draw=black, fill=yellow!30, drop shadow]
\tikzstyle{decision} = [diamond, rounded corners, minimum width=2.5cm, minimum height=1.6cm, text centered, draw=black, fill=green!30, drop shadow]
\tikzstyle{arrow} = [thick,->,>=stealth]
\tikzstyle{line} = [draw, -latex']
\begin{tikzpicture}[node distance=2.5cm and 2.5cm]

\node (start) at (0,0) [decision] {Problem};
\node (input1) at (1.5,1.75) [io] {Input: {\sc pddl}2.1};
\node (input2) at (1.5, -1.75) [io] {Input: General Description};

\node (PDDLconvert) at (4,1.75) [manual] {Translate {\sc pddl} to Rust};
\node (genconvert) at (4, -1.75) [manual] {Generate Rust Functions};

\node (merge) at (5.5, 0) [process] {Generate Heuristic Code via LLM};

\node (compile) at (8, 0)  [process] {Compile into GBFS and Run};

\node (verify1) at (9.5, 1.75) [process] {Verify with {\sc val}};
\node (verify2) at (9.5, -1.75) [process] {Verify with Custom Procedure};
\node (end) at (11, 0) [startstop] {Solution};

\draw [arrow] (start.north) |- (input1.west);
\draw [arrow] (start.south) |- (input2.west);

\draw [arrow] (input1.east) -- (PDDLconvert);
\draw [arrow] (input2.east) -- (genconvert);
\draw [arrow] (PDDLconvert.east) -| (merge.north);
\draw [arrow] (genconvert.east) -| (merge.south);

\draw [arrow] (merge) -- (compile);

\draw [arrow] (compile.north) |- (verify1.west);
\draw [arrow] (compile.south) |- (verify2.west);

\draw [arrow] (verify1.east) -| (end);
\draw [arrow] (verify2.east) -| (end);

\end{tikzpicture}
\caption{Procedure flowchart. A general problem description is manually written using Rust components---successor generator, goal test, and initial state---which are then fed into the system. For problems already described in \textsc{pddl}2.1, the Rust translation is derived directly from the encoding.}
\label{fig:algo-flow-horizontal}
\end{figure*}

\section{Methodology}
\label{section:method}
We start with the \textbf{overview} of our approach.  The process begins by accepting two types of input: structured representations provided in {\sc pddl}2.1 \citep{fox-long:jair-2003} format and general problem descriptions given in natural language or in other non-standardized formats. The problems specified in {\sc pddl}2.1 are manually translated into Rust functions that implement a \textit{successor generator} and a \textit{goal-testing} function. For general problem descriptions, we encode them into Rust according to their specifications. Moreover, the initial and the goal states are specified separately in a JSON file. The manual translation is necessary because current LLMs are not sufficiently reliable for this task, and existing tools lack the maturity required to perform it effectively. The automatic translations to ESG formalism is left for future work.

With the foundational components in place, we employ an LLM to generate the heuristic function. The LLM is provided with the Rust code for the successor generator and goal-testing function, along with a prompt designed to produce a heuristic function in Rust suited to the problem. 
The prompt\footnote{The exact prompt for both variations is given in the Appendix.} has three components:
\begin{enumerate}
\item Conditioning the model to be a senior engineer in the GPL used \cite{Anam2025PromptEA} and providing the format the resulting heuristic must follow. This component is spread across both the system prompt and the user message\footnote{We include guidance to avoid compilation errors, however the guidance could be \textit{far} better optimized. This optimization, however, is out of scope for this work.}
\item Requesting that the LLM generate a heuristic.
\item Providing the model with the GPL domain implementation (the ESG).
\end{enumerate}

As an additional variation of our method, the user message may also contain a fourth component:

\begin{enumerate}
\setcounter{enumi}{3}
\item The JSON representation of the instance to be solved.
\end{enumerate}

We analyze the impact of instance-specificity in the heuristic generation pipeline in the Ablation Study section.

The heuristic, successor generator, and goal-testing functions are then integrated into a standard GBFS framework, where the search is guided by the generated heuristic. To ensure correctness, all solutions are verified using appropriate mechanisms based on the input type. For problems specified in {\sc pddl}2.1, we rely on a standard {\sc pddl} verifier---{\sc val}~\citep{howey-etal:ictai-2004}, which cross-checks the solution against the formal specification of the problem. For general problem descriptions, we verify by applying the actions in the plan sequentially, verify the preconditions are met, and that the last state fulfills goal-test. For the visualization of the workflow see Fig.~\ref{fig:algo-flow-horizontal}. This workflow integrates LLM capabilities with automated code generation and classical search techniques, enabling efficient and flexible problem-solving for a wide range of planning domains. Note that the only element of the workflow that still requires human intervention is the specification of the problem.

\newcommand{\best}[1]{\cellcolor{blue!30}{\textbf{#1}}}
\newcommand{\second}[1]{\cellcolor{blue!15}{#1}}

\begin{table*}[!t]
\centering
\scriptsize
\setlength{\tabcolsep}{3.2pt}
\newcommand{\numtasks}[1]{\hfill (#1)}
\begin{tabular}{@{}lr|rr|rrrrr|rrr|rrr|rrr@{}}
 & & \multicolumn{2}{|c}{Baseline} & \multicolumn{5}{|c}{Domain-Independent Planners} & \multicolumn{3}{|c}{\textsc{FirstCompilation (fc)}} & \multicolumn{3}{|c}{\textsc{UntilSuccess (US)}} & \multicolumn{3}{|c}{\textsc{SelfPortfolio-10 (SP-10)}}\\
& & \multicolumn{2}{c|}{Rust} & \multicolumn{3}{c}{ENHSP-20} & \multicolumn{1}{c}{NFD} & \multicolumn{1}{c|}{MFF} &
\multicolumn{2}{c}{GPT} &
\multicolumn{1}{c|}{Claude} &
\multicolumn{2}{c}{GPT} &
\multicolumn{1}{c|}{Claude}  &
\multicolumn{2}{c}{GPT} &
\multicolumn{1}{c}{Claude}\\
Domain & & BFS & $h^{\text{md}}$ & $h^{\text{md}}$ & $h^{\text{add}}_{\langle \text{B, QB}\rangle}$ & $P(3h||3n)^{\ddagger}$ & $h_{\text{LMC}}$ & $h_{\text{FF}}$ & 4.1 & 5.1$^{\dagger}$ & Sonnet 4.5$^{\dagger}$ & 4.1 & 5.1$^{\dagger}$ & Sonnet 4.5$^{\dagger}$ & 4.1 & 5.1$^{\dagger}$ & Sonnet 4.5$^{\dagger}$\\
\midrule

Block Grouping & \numtasks{20} & 0 & 9 & 15 & 18 & \best{20}$^{\ddagger}$ & 0 & 2 & 16 & 11 & 15 & \second{19} & 17 & 18 & \second{19} & 18 & 18 \\
Counters & \numtasks{20}       & 3 & 10 & 13 & 11 & \best{20}$^{\ddagger}$ & 12 & \second{15} & 10 & 10 & 5 & 10 & 10 & 5 & 10 & 10 & 5 \\
Delivery & \numtasks{20}       & 1 & 9 & 17 & 12 & \second{18}$^{\ddagger}$ & 9 & \best{20} & 16 & 16 & 16 & \best{20} & 16 & \best{20} & \best{20} & 17 & \best{20} \\
Drone & \numtasks{20}          & 3 & 16 & \best{20} & 15 & \best{20}$^{\ddagger}$ & 16 & 16 & 15 & 16 & 16 & 16 & 16 & \second{17} & 16 & 16 & \best{20} \\
Expedition & \numtasks{20}     & 3 & \second{5} & 3 & \best{6} & \best{6}$^{\ddagger}$ & \second{5} & \second{5} & 3 & 3 & 2 & 3 & 3 & 4 & 3 & 3 & 4 \\
Farming & \numtasks{20}        & 3 & \best{20} & \best{20} & \best{20} & \best{20}$^{\ddagger}$ & 15 & 9 & 18 & \best{20} & \best{20} & \second{19} & \best{20} & \best{20} & \second{19} & \best{20} & \best{20} \\
FO-Counters & \numtasks{20}    & 3 & 7 & 9 & \second{15} & \second{15}$^{\ddagger}$ & 6 & \best{20} & 4 & 5 & 6 & 4 & 6 & 6 & 4 & 6 & 6 \\
FO-Farming & \numtasks{20}     & 3 & \best{20} & \best{20} & \best{20} & \best{20}$^{\ddagger}$ & 11 & 16 & \best{20} & \second{17} & \best{20} & \best{20} & \best{20} & \best{20} & \best{20} & \best{20} & \best{20} \\
FO-Sailing & \numtasks{20}     & 0 & 0 & 0 & 1 & 3$^{\ddagger}$ & 16 & 11 & \best{20} & 0 & 11 & \best{20} & \best{20} & \second{18} & \best{20} & \best{20} & \second{18} \\
Hydropower & \numtasks{20}     & 8 & 4 & 4 & \best{20} & \best{20}$^{\ddagger}$ & 6 & 1 & 8 & 6 & 8 & \second{12} & \second{12} & 8 & \second{12} & \best{20} & 8 \\
Market Trader & \numtasks{20}  & 0 & 7 & \second{17} & \best{20} & \best{20}$^{\ddagger}$ & 0 & 0 & \best{20} & 0 & 0 & \best{20} & 1 & \best{20} & \best{20} & 1 & \best{20} \\
Pathways & \numtasks{20}       & 0 & 1 & 0 & \second{2} & \best{13}$^{\ddagger}$ & \second{2} & \best{13} & 0 & 0 & 0 & 1 & 1 & 1 & 1 & 1 & 1 \\
Plant Watering & \numtasks{20} & 0 & 1 & \best{20} & \second{19} & \best{20}$^{\ddagger}$ & \best{20} & 13 & 9 & 10 & 6 & \best{20} & \best{20} & \best{20} & \best{20} & \second{19} & \best{20} \\
Rover & \numtasks{20}          & 2 & 4 & \second{10} & 6 & \best{14}$^{\ddagger}$ & 4 & \second{10} & 3 & 4 & 4 & 4 & 4 & 4 & 4 & 4 & 4 \\
Sailing & \numtasks{20}        & 0 & 0 & 0 & 17 & \best{20}$^{\ddagger}$ & 10 & 2 & 11 & \second{19} & \second{19} & \best{20} & \best{20} & \second{19} & \best{20} & \best{20} & \second{19} \\
Settlers & \numtasks{20}       & 0 & 1 & 0 & 0 & \best{8}$^{\ddagger}$ & 0 & \second{6} & 1 & 1 & 3 & 1 & 1 & 3 & 3 & 1 & 3 \\
TPP & \numtasks{20}            & 1 & \second{16} & \best{20} & 4 & \best{20}$^{\ddagger}$ & 2 & 4 & 13 & 0 & 0 & \second{16} & 4 & 2 & \second{16} & 4 & 2 \\
Zenotravel & \numtasks{20}     & 4 & 11 & 14 & 11 & \second{18}$^{\ddagger}$ & 10 & 0 & \best{20} & \best{20} & 14 & \best{20} & \best{20} & 17 & \best{20} & \best{20} & 17 \\
\midrule
$\sum$ & \numtasks{360}        & 34 & 141 & 202 & 217 & \best{295}$^{\ddagger}$ & 144 & 163 & 207 & 158 & 165 & 245 & 211 & 222 & \second{247} & 220 & 225 \\
\midrule
Twin Prime & \numtasks{20}     & 4 & \best{20} & - & - & - & - & - & \second{18} & 17 & 17 & \second{18} & 17 & 17 & \second{18} & \best{20} & 17 \\
Pacman & \numtasks{20}         & 1 & \second{14} & - & - & - & - & - & 9 & 13 & 10 & 9 & \second{14} & \best{16} & \best{16} & \best{16} & \best{16} \\
\midrule
$\sum$ & \numtasks{400}        & 39 & 175 & - & - & - & - & - & 234 & 188 & 192 & \second{272} & 242 & 255 & \best{281} & 256 & 258 \\

\end{tabular}
    \caption{\label{table-coverage} Coverage results of the baseline, domain-independent planners, the best portfolio, and the LLM-generated heuristics. We report our results from GPT-4.1, GPT-5.1, and Claude Sonnet 4.5 with configurations \textsc{FirstCompilation (fc)}, \textsc{UntilSuccess (us)},  and \textsc{SelfPortfolio-10 (SP-10)}. All configuration except BFS and MFF (which uses EHCS) use GBFS. Models marked with $^{\dagger}$ are reasoning models. For results on all models and configurations see Appendix. $^{\ddagger}$ The $P(3h||3n)$ portfolio uniformly divides time among three “best’’ ENHSP heuristics and three “best’’ novelty heuristics \cite{chen-thiebaux:socs-2024}, where the notion of “best’’ is derived from performance on the same problem set used in our evaluation, rather than being fixed independently of it. }
\end{table*}

\section{Empirical Evaluation}
\label{section:evaluation}

The experiments were conducted on a 13th Gen Intel® Core™ i9-13900 processor running at 2.00 GHz, supported by a 64-bit operating system and 32 GB of RAM. To ensure consistency and reliability, each experiment was constrained by a time limit of 10 minutes and a memory usage limit of 8 GB. We test our approach in the context of satisficing numeric planning. We have evaluated our approach on the domains from the Numeric International Planning Competition (IPC) 2023 \citep{taitler-etal:aimag-2024}.\footnote{https://github.com/ipc2023-numeric} Note that even problems with three unbounded numeric variables have infinite size state spaces and in general are undecidable \cite{gnad-et-al:icaps-2023}. As a proof-of-concept, we also include two domains that cannot be easily expressed using {\sc pddl}:\\
\noindent\textbf{Twin Prime:} The initial state comprises a set of integers and buffers, each holding a single integer. The actions include addition, subtraction, multiplication, and integer division between two registers, with results stored in one register. The goal is to produce a twin prime number exceeding a specified threshold in one of the buffers---a goal that is not easily expressible in {\sc pddl}.

\noindent\textbf{Deterministic Pacman:} The game is defined on a grid containing walls, pellets, ghosts, power-ups, and a single Pacman agent. Pacman moves in one of four cardinal directions per turn, consuming pellets and power-ups upon entry into their cell. The goal is to clear all pellets while avoiding collisions with ghosts that behave in a deterministic manner. Consuming a power-up gives Pacman a temporary ability to banish ghosts and avoid harm from them. This domain cannot be easily expressed in {\sc pddl} or its extensions due to the complex interactions between Pacman, ghosts, and power-ups, which require temporal dynamics and conditional effects beyond {\sc pddl}’s representational capabilities.


\begin{figure*}[thp]
    \centering

    \begin{subfigure}[b]{0.8\textwidth}
        \centering
        \includegraphics[width=\textwidth]{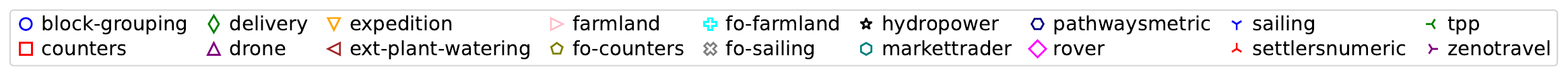}
    \end{subfigure}

    \begin{subfigure}[b]{0.48\textwidth}
        \centering
        \includegraphics[width=\textwidth]{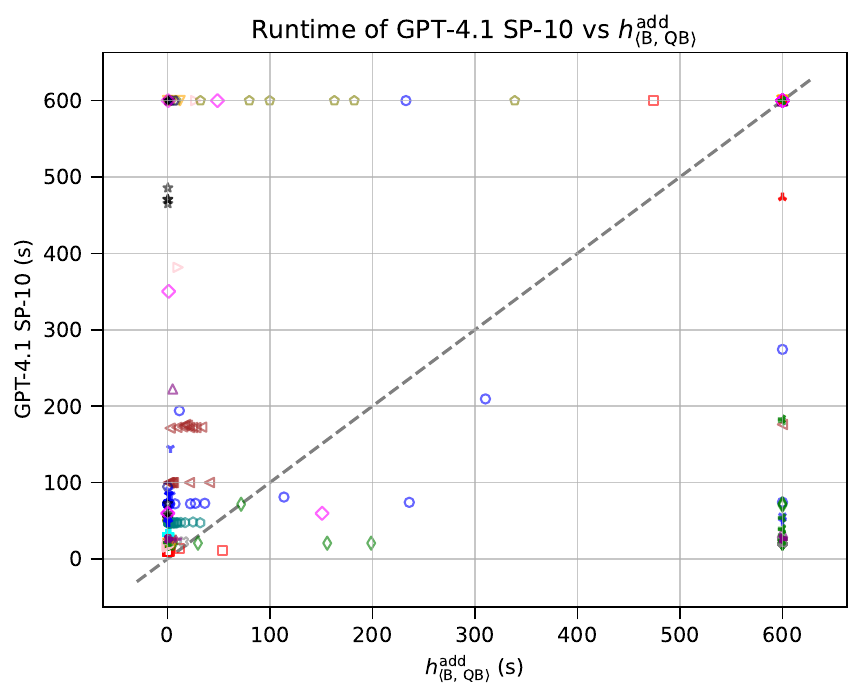}
        \caption{Total Time (including failed generation + run pairs)}
    \end{subfigure}
    \hfill
    \begin{subfigure}[b]{0.48\textwidth}
        \centering
        \includegraphics[width=\textwidth]{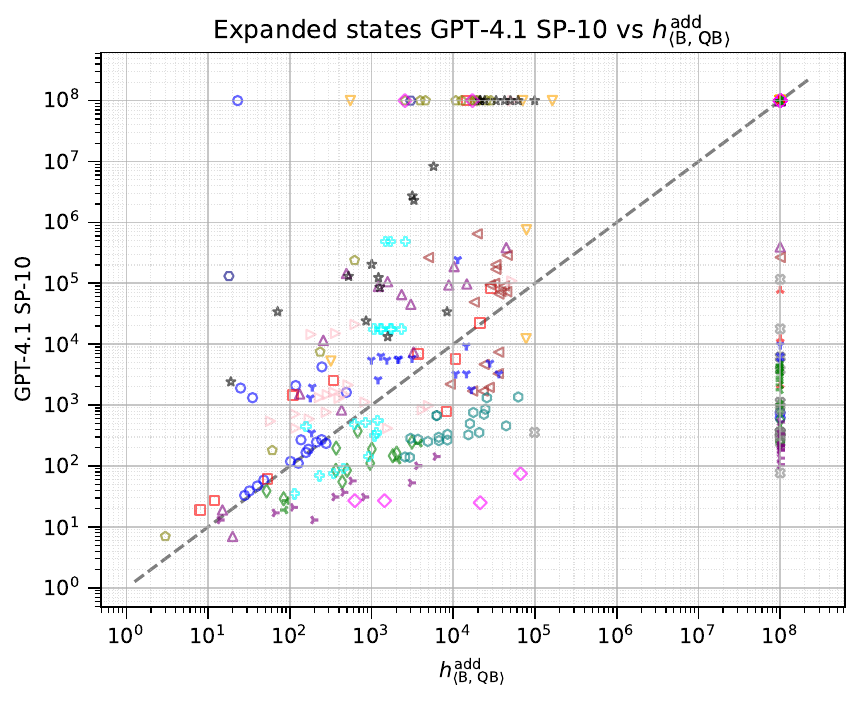}
        \caption{Expanded States}
    \end{subfigure}

    \caption{Per-instance comparisons of the Total Time (left) and expanded states (right) between GPT-4.1 with SelfPortfolio-10 and $h^{\text{add}}_{\langle \text{B, QB}\rangle}$. Points below the diagonal favor our approach. On problems both solve they seem to provide similar levels of heuristic guidance.\footnotemark}
    \label{fig:SP10_vs_novelty}
\end{figure*}

\footnotetext{A portfolio between the two achieves a coverage of 304, however this is a post-hoc optimization and not a fair comparison. We do however consider Domain-Independent+LLM a fruitful direction fo future work.}

\begin{figure}[ht]
    \centering
    \includegraphics[width=0.47\textwidth]{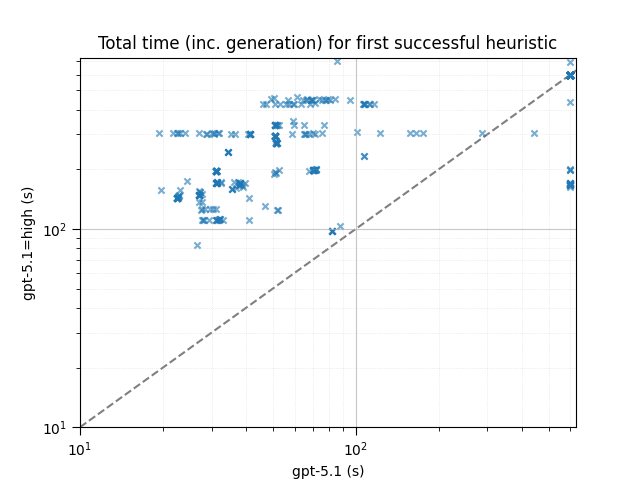}%
    \hfill
    \includegraphics[width=0.47\textwidth]{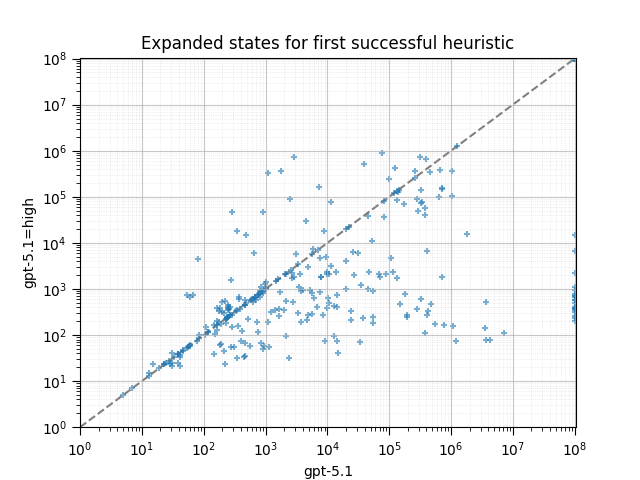}
    \caption{\label{fig:ab}Per-instance comparison of the Total Time (generation + run) (up) and expanded states (down) between GPT-5.1 with and without setting the reasoning\_effort to \textit{high}. Points below the diagonal favor high reasoning effort. Allowing increased reasoning effort moderately increases heuristic quality, but comes at a heavy expense of time.}
\end{figure}

We evaluated our approach against the following state-of-the-art methods on the IPC domains. All of the planners below support linear expressions in both conditions and assignment effects. As a \textit{baseline}, we implement a ``blind'' BFS and GBFS with a manually implemented $h^{\text{md}}$ heuristic~\cite{chen-thiebaux:socs-2024}, running on a Rust implementation of the problems based on \cite{Domains_Project}.


The $h^{\text{md}}$ heuristic estimates the total distance from a state $s$ to satisfying all goal conditions, inspired by the $h^{\text{md}}$ norm:
\[
h^{\text{md}}(s) := \sum_{\varepsilon \in G} d(\varepsilon, s),
\]
where $G$ is the set of goal conditions, and $d(\varepsilon, s)$ measures the distance from $s$ to satisfying $\varepsilon$.

For propositional goals, $d(\varepsilon, s) = 0$ if $s \models \varepsilon$ and $1$ otherwise.  
For numeric goals of the form $\varepsilon: \psi \,\bowtie\, 0$, where $\psi: \mathbb{R}^n \to \mathbb{R}$ and $\bowtie \, \in \{>, \geq, =, \leq, <\}$, we define:
\[
d(\varepsilon, s) := \inf_{y : \psi(y) \,\bowtie\, 0} |\psi(y) - \psi(s)|,
\]
i.e., the distance from $\psi(s)$ to the nearest satisfying value of $\psi$.  
\citet{chen-thiebaux:socs-2024} proposed this approach for linear constraints in {\sc pddl}2.1, but this heuristic extends naturally to non-linear formulas, though computing $d$ becomes more expensive.

\noindent\textbf{ENHSP-20} is a Java-based planner whose satisficing configurations use GBFS by \citep{scala-et-al:jair-2020}, using the modified version released by \citet{chen-thiebaux:socs-2024}.\footnote{\url{github.com/hstairs/jpddlplus/tree/socs24-width-and-mq}}. Following \citep{chen-thiebaux:socs-2024} we compare against $h^{\text{md}}$, $h^{\text{mrp}}_{\text{+hj}}$, $h^{\text{add}}$, their novelty variants  $h^{\text{md}}_{\langle \text{B, QB}\rangle}$, $h^{\text{mrp}}_{\text{+hj}~\langle \text{B, QB}\rangle}$, $h^{\text{add}}_{\langle \text{B, QB}\rangle}$, as well as the three portfolio settings, $P(3h)$ (non-novelty only), $P(3n)$ (novelty only), and $P(3h \,\|\, 3n)$ (both). 
We report of them only $h^{\text{md}}$, the best-performing single heuristic $h^{\text{add}}_{\langle \text{B, QB}\rangle}$, and the best portfolio configuration, $P(3h \,\|\, 3n)$, while the full comparison is provided in Appendix.

\noindent\textbf{Metric-FF (MFF)} \citep{hoffman:jair-2003} is used off-the-shelf. Implemented in C, it employs Enforced Hill Climbing Search (EHCS) with the interval-relaxed $h_{\text{FF}}$ heuristic and Helpful Operators.\footnote{fai.cs.uni-saarland.de/hoffmann/metric-ff.html} The planner is incomplete and may report problems as unsolvable due to limitations of EHCS.\\
\noindent\textbf{Numeric-FD (NFD)} \citep{aldinger-nebel:socs-2017} is a C++ planner that employs Lazy GBFS with numeric LM-cut heuristics \citep{kuroiwa-et-al:icaps-2022}.\footnote{github.com/ipc2023-numeric/team-1} This planner won the Numeric IPC-2023 \cite[see][]{taitler-etal:aimag-2024}.

The \textbf{Pacman} and \textbf{Twin Prime} domains demand a level of expressiveness that makes encoding them for existing planners highly impractical and cumbersome in practice. As there currently are no available methods to derive a domain-independent heuristic for this type of tasks, we benchmarked them against BFS and $h^{\text{md}}$.

LLM-generated heuristics are not always compilable and vary in quality and efficiency generation-to-generation. To address this, we employ three fallback strategies:

In \textsc{FirstCompilation (fc)}, we repeatedly query the LLM until a compilable heuristic is obtained and run GBFS with it until a solution is found or resources are exhausted, 600 sec (search + all API calls).

In \textsc{UntilSuccess (us)}, we operate the same as \textsc{fc}, except we continue even if a compilable heuristic failed.

In \textsc{SelfPortfolio-N (sp-n)}, we allocate fixed $Time/N$s slices to N heuristics; if a run fails or exceeds memory, we restart with generating a new heuristic. This continues until a solution is found or all slices failed. All API call durations, typically 10–45 seconds depending on the model, are included in the time budget.

We evaluated the following LLMs and settings: OpenAI’s GPT-4.1 and GPT-4.1-mini, their reasoning models GPT-5.1 and GPT-5-mini (under both a "low" and "high" reasoning effort setting), and Anthropic’s Claude Sonnet 4.5 and Claude Haiku 4.5. Results are presented in Table \ref{table-coverage}. For practicality, we limit the maximum number of heuristics generated for \textsc{fc} and \textsc{us} to 10, which if passed the instance is declared failed.
As is standard in AI planning, we report coverage rather than averages, since state space size typically grows exponentially and averaging would be misleading. As a qualitative measure of the generated heuristics we report the head-to-head time and expanded nodes for solution comparison of \textsc{sp-10} with GPT-4.1 vs. the best performing $h^{\text{add}}_{\langle \text{B, QB}\rangle}$, which can be seen in Figure~\ref{fig:SP10_vs_novelty}. Comparisons with additional models can be seen in Appendix.

\begin{figure*}[ht]
    \centering
    \includegraphics[width=1.0\textwidth]{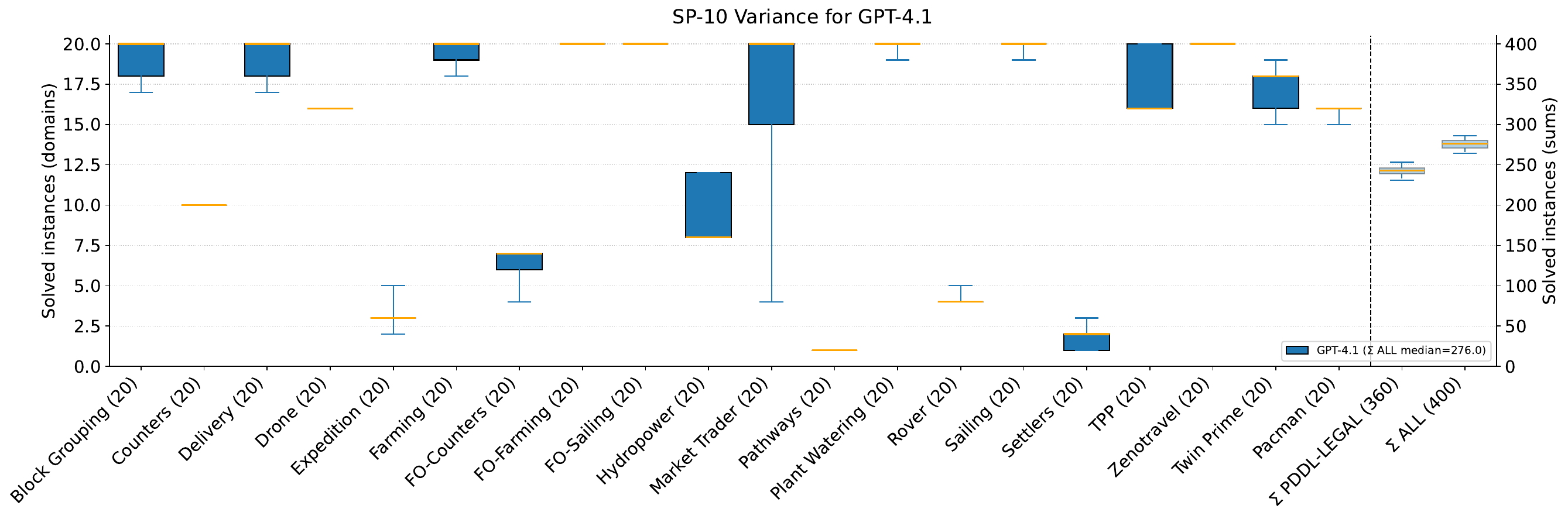}
    \caption{As a variance analysis we generated 40 heuristics with GPT-4.1 in each domain, and performed a Monte-Carlo simulation of our algorithm for 1000 iterations of sampling heuristic order, taking the coverage per-domain and overall each time. The boxes represent the first and third quartiles, the thick orange line the median, and the whiskers the 1st and 99th percentiles. Although limited, the analysis shows the method to be highly consistent for most domains.}
    \label{fig:variance}
\end{figure*}

We conducted ablation studies to evaluate the contribution of three components in the heuristic generation pipeline: (i) the provision of the specific instance (ii) the model and fallback choice (iii) the reasoning effort requested from reasoning models. The study evaluated their effect on heuristic quality, generation time, planning performance, and the variability of our best configuration (Table \ref{fig:variance}).

\paragraph{Impact of Instance-Specificity} 
We evaluate the impact of providing the instance JSON, which contains the initial state and goal, before generation. This incurs a significant cost, as it requires generating heuristics per instance, unlike the \textit{domain-specific-instance-general} approach, which is generated once and applied to all instances. Intuitively, IS may offer the following benefits:
\begin{compactenum}
\item The heuristic often contains multiple penalizing elements which are combined with guessed coefficients. Knowing the details of the instance gives the model a better idea on how to tune these coefficients.
\item It tells the model how significant each part of the problem is. Consider a simple instance of Pacman where the ghosts are trapped. The model can avoid spending generation- and run-time on a complex subroutine to quantify the risk from the ghosts. 
\item It tells the model the scale of the problem which can provide guidance on efficiency constraints. We hypothesize that there exists a tradeoff between the precision of the heuristic estimates and the computation time, and further that for larger problems it can be worthwhile for the heuristic to sacrifice precision for efficiency. For example, on TPP, Anthropic's models implement path-finding algorithms like Dijkstra, Prim and Floyd-Warshall, about half the time. Although this is a good basis for a ``smart'' heuristic, it causes a timeout within a few thousand expansions for even medium problems. Meanwhile even Manhattan Distance heuristics achieves 16/20.
\item At times, a solution state is obvious from the initial state (e.g, FO-Counters). In these cases showing the model the instance details allows it to write a heuristic to guide efficiently toward that solution rather than abstractly guiding toward all solutions.
\end{compactenum}

Empirically, IS leads to a decrease in overall coverage (Table~\ref{tab:refine-ablation}) with most domains varying slightly but a few increasing or decreasing significantly. IS did not cause heuristic generation time to increase\footnote{Interestingly, despite adding tokens and theoretically adding complexity which can increase output tokens, adding the instance slightly \textit{decreased} generation time. (P=2e-5$<$0.05). We do not have an explanation for this fact.}. IS moderately increased compilation errors (see Table ~\ref{tab:costs-table}), and if this effect is normalized against, IS-heuristics have a slightly higher solving rate. We hypothesize that GPT-4.1 is insufficiently strong to make use of this info and that for a stronger model it may be worth revisiting.

\begin{table}[!t]
\centering
\setlength{\tabcolsep}{3.2pt}
\newcommand{\numtasks}[1]{\hfill (#1)}
\begin{tabular}{@{}lr|rr|rr|rr@{}}
 & & \multicolumn{2}{|c}{FC} & \multicolumn{2}{|c}{US} & \multicolumn{2}{|c}{SP-10}\\
Domain & & DD & IS & DD & IS & DD & IS\\
\midrule
Block Grouping &\numtasks{20}    & 16  &	9   & 19  & 10  & \textbf{19}  & 10  \\
Counters &\numtasks{20}          & 10  &	8   & 10  & 11  & 10  & \textbf{11}  \\
Delivery &\numtasks{20}          & 16  &	10  & 20  & 18  & \textbf{20}  & 18  \\
Drone &\numtasks{20}             & 15  &	15  & 16  & 16  & \textbf{16}  & \textbf{16}  \\
Expedition &\numtasks{20}        & 3   &	1   & 3   & 4   & 3 \textbf{}  & \textbf{4}   \\
Farming &\numtasks{20}           & 18  &	19  & 19  & 20  & 19  & \textbf{20}  \\
FO-Counters& \numtasks{20}       & 4   &	8   & 4   & 14  & 4   & \textbf{14}  \\
FO-Farming& \numtasks{20}        & 20  &	18  & 20  & 20  & \textbf{20}  & \textbf{20}  \\
FO-Sailing& \numtasks{20}        & 20  &	10  & 20  & 19  & \textbf{20}  & \textbf{20}  \\
Hydropower& \numtasks{20}        & 8   &	7   & 12  & 9   & \textbf{12}  & 9   \\
Market Trader& \numtasks{20}     & 20  &	1   & 20  & 3   & \textbf{20}  & 3   \\
Pathways& \numtasks{20}          & 0   &	1   & 1   & 1   & \textbf{1}   & \textbf{1}   \\
Plant Watering &\numtasks{20}    & 9   &	12  & 20  & 18  & \textbf{20}  & 18  \\
Rover& \numtasks{20}             & 3   &	2   & 4   & 3   & \textbf{4}   & 3   \\
Sailing &\numtasks{20}           & 11  &	14  & 20  & 20  & \textbf{20}  & \textbf{20}  \\
Settlers& \numtasks{20}          & 1   &	2   & 1   & 3   & \textbf{3}   & \textbf{3}   \\
TPP& \numtasks{20}               & 13  &	11  & 16  & 13  & \textbf{16}  & 15  \\
Zenotravel& \numtasks{20}        & 20  &	14  & 20  & 20  & \textbf{20}  & \textbf{20}  \\
\midrule
$\sum$ &\numtasks{360}           & 207 &	162 & 245 & 222 & \textbf{247} & 225 \\
\midrule
Twin Prime& \numtasks{20}        & 18  &	20  & 18  & 20  & 18  & \textbf{20}  \\
Pacman &\numtasks{20}            & 9   &	13  & 9   & 13  & \textbf{16}  & 14  \\
\midrule
$\sum$ &\numtasks{400}           & 234 &	195 & 272 & 255 & \textbf{281} & 259 \\
\end{tabular}
\caption{ Coverage results for GPT-4.1 with domain-dependence only (DD), our standard configuration, against GPT-4.1 with Instance-Specificity (IS), reporting each on \textsc{fc}, \textsc{us}, and their best configuration, SP-10. For most domains GPT-4.1 fails to leverage the additional information, however we highlight FO-Counters and Twin Prime as domains where it lead to the most improvement.}
\label{tab:refine-ablation}
\end{table}

\paragraph{Impact of model and fallback choice}

We tested 6 models: GPT-4.1, GPT-4.1-mini, GPT-5.1, GPT-5-mini, Claude Sonnet 4.5, and Claude Haiku 4.5, each with each of our fallback options including \textsc{fc}, \textsc{us}, \textsc{sp-5}, and \textsc{sp-10}. For the full table see Appendix. Our observations are as follows: (i) for non-reasoning models, under about every configuration, a model's larger variant is preferred. Since they are well-capable of generating and running many heuristics within the allotted time, $SP-10$ is preferred; (ii) reasoning models are more competitive between sizes. They take longer to generate and $US$ performs best, allowing them the time they need; (iii) anecdotally, when reasoning models fail, it is often because they overcomplicate the problem, leading to longer generation times and unpredictable outcomes---sometimes producing better heuristics, other times unusable ones. That said, this risk may be acceptable if we can resample and the samples are sufficiently independent.

\paragraph{Impact of Reasoning effort}

We tested our method on the same model, GPT-5.1 with two configurations: \textit{reasoning\_effort='low'} and \textit{reasoning\_effort='high'} (for brevity denoted 'GPT-5.1' and 'GPT-5.1=high'). We found that although increasing $reasoning\_effort$ slightly increased heuristic efficiency and quality, and better-than-halved compilation error rate, when we account for generation time the reduced number of heuristics we generate in time does not make this trade worthwhile. With better engineering one could generate many completions simultaneously, which may turn this trade worthwhile. We leave this direction for future work. For further graphs see appendix.

\begin{table*}[ht!]
    \centering
    \begin{tabular}{l||ccccccc|cc}
        \toprule
        & \multicolumn{7}{c}{\textbf{GPT}} & \multicolumn{2}{c}{\textbf{Claude}}\\
        \cmidrule(lr){2-8} \cmidrule(lr){9-10}
        Metric
        & \textbf{4.1}
        & \textbf{4.1 (IS)}
        & \textbf{4.1 mini}
        & \textbf{5.1$^{\dagger}$}
        & \textbf{5 mini$^{\dagger}$}
        & \textbf{5.1 high$^{\dagger}$}
        & \textbf{5 m-h$^{\dagger}$}
        & \textbf{Sonnet 4.5$^{\dagger}$}
        & \textbf{Haiku 4.5$^{\dagger}$} \\
        \midrule
        Input tok.           & 3725 & 8102 & 3725 & 3724 & 3724 & 3724  & 3724  & 4761 & 4761  \\
        Output tok.          & 1510 & 1695 & 1245 & 3253 & 2298 & 18075 & 14485 & 2122 & 1490  \\
        Reasoning tok.\footnotemark       & -    & -    & -    & 1282 & 589  & 15510 & 11613 & -    & -     \\
        Cost (USD)             & 0.02 & 0.03 & 0.003& 0.037& 0.006& 0.185 & 0.03  & 0.046& 0.012 \\
        Generation (sec.)    & 35.8 & 27.5 & 19.4 & 44.9 & 44.5 & 235.1 & 202.3 & 33.6 & 12.5  \\
        Run (sec.)           & 24.7 & 22.1 & 19.2 & 24.2 & 18.0 & 21.4  & 36.7  & 43.4 & 22.2  \\
        \midrule\midrule
        Compl. error (rate)  & 0.300& 0.430& 0.215& 0.150& 0.405& 0.070 & 0.205 & 0.195& 0.165 \\
        OOM (rate)               & 0.304& 0.241& 0.388& 0.382& 0.280& 0.340 & 0.288 & 0.391& 0.451 \\
        Runtime error (rate)    & 0.027& 0.020& 0.035& 0.032& 0.028& 0.035 & 0.032 & 0.016& 0.032 \\
        Timeout (rate)          & 0.015& 0.014& 0.004& 0.007& 0.009& 0.007 & 0.031 & 0.032& 0.002 \\
        Success (rate)           & 0.355& 0.295& 0.358& 0.430& 0.278& 0.548 & 0.444 & 0.365& 0.351 \\
        \bottomrule
    \end{tabular}
    \caption{Average token usage, cost, latency, and execution outcomes for different GPT and Claude models, independent of downstream applications. Models marked with $^{\dagger}$ are reasoning models. Excepting IS, input tokens across models depend only on the tokenizer. Costs do not account for token caching discounts.}
    \label{tab:costs-table}
\end{table*}

\footnotetext{For GPT, subset of output tokens. For Claude, they are not provided by the API}

The ablation results suggest that non-reasoning models don't significantly benefit from IS. The best setting of our method is to use a non-reasoning model to generate non-IS heuristics and run them as SelfPortfolio-10 (276 with GPT-4.1). If time is allowed ahead of receiving the instances, the best setting becomes instead to generate ahead of time as many heuristics as possible and run SP-N or US on them (The two fallbacks are approximately equivalent when used on pregenerated heuristics due to near-zero timeout rate).

\section{Discussion}

As shown in Table \ref{table-coverage}, LLM-based heuristics achieve state-of-the-art performance across the evaluated domains. They are particularly effective in challenging benchmarks such as \textbf{(FO-) Sailing} and \textbf{Zenotravel}, which are known to be difficult for domain-independent planners.
The best-performing LLM-based heuristics were produced by GPT-5.1 and it's mini variant with high reasoning effort, however when accounting for generation time and cost, GPT-4.1 emerges as a clear winner. Nonetheless, certain domains remain challenging for LLMs, likely due to their sensitivity to non-descriptive variable names and opaque goal specifications, as seen in \textbf{Pathways} and \textbf{Settlers}.
\paragraph{Expressiveness} In the \textbf{Pacman} and \textbf{Twin Prime} domains, no available planner can solve problems requiring this level of expressive power, necessitating a comparison against BFS and $h^{\text{md}}$. As expected, LLM-based heuristics outperformed both. Overall, LLM-based heuristics surpass state-of-the-art numerical planners while addressing problems that cannot be adequately expressed in any existing {\sc pddl} dialect.
\paragraph{Representation} One concern is that the performance gains observed with LLM-generated heuristics may result from optimized planner code rather than the heuristics themselves. To test this, we reimplemented $h^{\text{md}}$ within our planner for a direct comparison with ENHSP-20 using the same heuristic. As shown in Table \ref{table-coverage}, ENHSP-20 with $h^{\text{md}}$ generally achieves better results, particularly in domains such as \textbf{Hydropower} and \textbf{Plant Watering}, which are highly sensitive to problem representation. This suggests that departing from {\sc pddl}-based input to a direct successor-generator representation may, in some cases, degrade performance, allowing us to reject the hypothesis that our representation alone accounts for the observed improvements.

\noindent\textbf{Costs and Efficiency} Although concerns about LLM efficiency have been raised \citep{stojkovic2024towards}, energy use is beyond this work’s scope. Our approach supports efficient deployment: heuristics can be generated once per domain and reused across instances, reducing cost even with powerful models. The resulting heuristics are also fast to run and tend to succeed or fail more quickly than domain-independent ones (see Appendix).

\section{Conclusion}
\label{section:conclusion}
This paper explored the use of LLMs to generate heuristic functions directly from AI planning task definitions, bypassing traditional domain-independent heuristics written in formal languages like {\sc pddl}. Our approach uses general-purpose code (Rust) to represent planning tasks and lets LLMs synthesize tailored heuristics.

Empirically, the results show that LLM-generated heuristics achieve state-of-the-art performance across several established planning benchmarks. The approach excels in domains traditionally difficult for domain-independent planners, such as \textbf{Zenotravel}, indicating substantial potential for efficiency gains. Furthermore, it successfully addresses complex planning tasks that are not easily expressible in conventional formalisms, like \textbf{Pacman} and \textbf{Twin Prime}.

Nonetheless, limitations remain. LLM-generated heuristics are sensitive to the clarity and descriptiveness of task representations; unintuitive state-space encodings degrade their quality. Moreover, while larger models such as GPT-4.1 and Sonnet 4.5 tend to produce high-quality heuristics, they incur greater computational costs. Smaller models offer improved cost-efficiency but often generate less reliable code. This trade-off highlights the importance of selecting models according to task complexity and scalability requirements.

This work opens several promising avenues for future research. These include improving robustness across domains, comparing LLM-generated heuristics to hand-crafted ones, and exploring hybrid strategies combining LLMs with domain knowledge or iterative refinement techniques.

\section*{Acknowledgements}

 Alexander Shleyfman’s and Yonatan Vernik's work was supported by ISF grant 2443/23.

\bibliography{aaai2026}

@article{bonet2001planning,
  title={Planning as heuristic search},
  author={Bonet, Blai and Geffner, H{\'e}ctor},
  journal={Artificial Intelligence},
  volume={129},
  number={1-2},
  pages={5--33},
  year={2001},
  publisher={Elsevier}
}

@article{liu2024evolution,
  title={Evolution of heuristics: Towards efficient automatic algorithm design using large language model},
  author={Liu, Fei and Tong, Xialiang and Yuan, Mingxuan and Lin, Xi and Luo, Fu and Wang, Zhenkun and Lu, Zhichao and Zhang, Qingfu},
  journal={arXiv preprint arXiv:2401.02051},
  year={2024}
}

@article{ye2024reevo,
  title={Reevo: Large language models as hyper-heuristics with reflective evolution},
  author={Ye, Haoran and Wang, Jiarui and Cao, Zhiguang and Berto, Federico and Hua, Chuanbo and Kim, Haeyeon and Park, Jinkyoo and Song, Guojie},
  journal={Advances in neural information processing systems},
  volume={37},
  pages={43571--43608},
  year={2024}
}

@article{correa2025classical,
  title={Classical Planning with LLM-Generated Heuristics: Challenging the State of the Art with Python Code},
  author={Corr{\^e}a, Augusto B and Pereira, Andr{\'e} G and Seipp, Jendrik},
  journal={arXiv preprint arXiv:2503.18809},
  year={2025}
}

@Article{helmert:jair-2006,
  author =       "Malte Helmert",
  title =        "The {Fast} {Downward} Planning System",
  journal =      "JAIR",
  year =         "2006",
  volume =       "26",
  pages =        "191--246"
}

@InProceedings{edelkamp:icapsws-2003,
  author =       "Stefan Edelkamp",
  title =        "Limits and Possibilities of {PDDL} for Model
                  Checking Software",
  booktitle =    "Proceedings of the {ICAPS 2003} Workshop on the
                  Competition: Impact, Organisation, Evaluation,
                  Benchmarks",
  year =         "2003"
}

@Article{fox-long:jair-2003,
  author =       "Maria Fox and Derek Long",
  title =        "{PDDL2.1}: {An} Extension to {PDDL} for Expressing
                  Temporal Planning Domains",
  journal =      "JAIR",
  year =         "2003",
  volume =       "20",
  pages =        "61--124"
}

@Book{russell-norvig:1995,
  author =       "Stuart Russell and Peter Norvig",
  title =        "Artificial {I}ntelligence --- {A} Modern Approach",
  publisher =    "Prentice Hall",
  year =         "1995"
}

@TechReport{edelkamp-hoffmann:tr-2004,
  author =       "Stefan Edelkamp and J{\"o}rg Hoffmann",
  title =        "{PDDL2.2}: {The} Language for the Classical Part of the
                  4th {International} {Planning} {Competition}",
  institution =  "Albert-Ludwigs-Universit{\"a}t Freiburg, Institut f{\"u}r
                  Informatik",
  number =       "195",
  year = 	 "2004"
}

@Book{pearl:1984,
  author =       "Judea Pearl",
  title =        "Heuristics: {Intelligent} Search Strategies for
                  Computer Problem Solving",
  publisher =    "Addison-Wesley",
  year =         "1984"
}

@Article{hart-et-al:ieee-ssc-1968,
  author =       "Peter E. Hart and Nils J. Nilsson and Bertram Raphael",
  title =        "A Formal Basis for the Heuristic Determination of
                  Minimum Cost Paths",
  journal =      "IEEE Transactions on Systems Science and Cybernetics",
  year =         "1968",
  volume =       "4",
  number =       "2",
  pages =        "100--107"
}

@Book{ghallab-et-al:2004,
  author =       "Malik Ghallab and Dana Nau and Paolo Traverso",
  title =        "Automated Planning: {Theory} and Practice",
  publisher =    "Morgan Kaufmann",
  year =         "2004"
}

@article{wei2022chain,
  title={Chain-of-thought prompting elicits reasoning in large language models},
  author={Wei, Jason and Wang, Xuezhi and Schuurmans, Dale and Bosma, Maarten and Xia, Fei and Chi, Ed and Le, Quoc V and Zhou, Denny and others},
  journal={Advances in neural information processing systems},
  volume={35},
  pages={24824--24837},
  year={2022}
}

@article{stojkovic2024towards,
  title={Towards Greener LLMs: Bringing Energy-Efficiency to the Forefront of LLM Inference},
  author={Stojkovic, Jovan and Choukse, Esha and Zhang, Chaojie and Goiri, Inigo and Torrellas, Josep},
  journal={arXiv preprint arXiv:2403.20306},
  year={2024}
}

@InProceedings{helmert-domshlak:icaps-2009,
  author =       "Malte Helmert and Carmel Domshlak",
  title =        "Landmarks, Critical Paths and Abstractions:
                  {What's} the Difference Anyway?",
  pages =        "162--169",
  booktitle =    "ICAPS",
  year =         "2009",
}

@TechReport{mcdermott-et-al:tr-1998,
  author =       "Drew McDermott and Malik Ghallab and Adele Howe
                  and Craig Knoblock and Ashwin Ram and Manuela Veloso
                  and Daniel Weld and David Wilkins",
  title =        "{PDDL} -- {The} {Planning} {Domain} {Definition}
                  {Language} -- {Version} 1.2",
  institution =  "Yale Center for Computational Vision and Control",
  year =         "1998",
  number =       "CVC TR-98-003"
}

@inproceedings{scala-et-al:ijcai-2016,
  author    = {Enrico Scala and
               Patrik Haslum and
               Sylvie Thi{\'{e}}baux},
  title     = {Heuristics for Numeric Planning via Subgoaling},
  booktitle = {IJCAI},
  pages     = {3228--3234},
  year      = {2016},
}

@article{scala-et-al:jair-2020,
  author    = {Enrico Scala and
               Patrik Haslum and
               Sylvie Thi{\'{e}}baux and
               Miquel Ram{\'{\i}}rez},
  title     = {Subgoaling Techniques for Satisficing and Optimal Numeric Planning},
  journal   = {JAIR},
  volume    = {68},
  pages     = {691--752},
  year      = {2020},
}

@inproceedings{aldinger-nebel:socs-2017,
  author    = {Johannes Aldinger and
               Bernhard Nebel},
  title     = {Interval Based Relaxation Heuristics for Numeric Planning with Action
               Costs},
  booktitle = {SOCS},
  pages     = {155--156},
  year      = {2017},
}

@article{hoffman:jair-2003,
  author    = {J{\"{o}}rg Hoffmann},
  title     = {The Metric-FF Planning System: Translating ''Ignoring Delete Lists'' to Numeric State Variables},
  journal   = {JAIR},
  volume    = {20},
  pages     = {291--341},
  year      = {2003},
}

@inproceedings{gnad-et-al:icaps-2023,
  author       = {Daniel Gnad and
                  Malte Helmert and
                  Peter Jonsson and
                  Alexander Shleyfman},
  title        = {Planning over Integers: Compilations and Undecidability},
  booktitle    = {ICAPS},
  pages        = {148--152},
  publisher    = {{AAAI} Press},
  year         = {2023},
}

@article{kuroiwa-et-al:jair-2022,
  author       = {Ryo Kuroiwa and
                  Alexander Shleyfman and
                  Chiara Piacentini and
                  Margarita P. Castro and
                  J. Christopher Beck},
  title        = {The LM-Cut Heuristic Family for Optimal Numeric Planning with Simple
                  Conditions},
  journal      = {J. Artif. Intell. Res.},
  volume       = {75},
  pages        = {1477--1548},
  year         = {2022}
}

@inproceedings{kuroiwa-et-al:icaps-2022,
  author    = {Ryo Kuroiwa and Alexander Shleyfman and J. Christopher Beck},
  title     = {LM-Cut Heuristics for Optimal Linear Numeric Planning},
  booktitle = {ICAPS},
  year      = {2022}
}

@inproceedings{howey-etal:ictai-2004,
  author       = {Richard Howey and
                  Derek Long and
                  Maria Fox},
  title        = {{VAL:} Automatic Plan Validation, Continuous Effects and Mixed Initiative
                  Planning Using {PDDL}},
  booktitle    = {ICTAI}, 
  pages        = {294--301},
  publisher    = {{IEEE} Computer Society},
  year         = {2004}
}

@inproceedings{guan-etal:nips-2023,
  author       = {Lin Guan and
                  Karthik Valmeekam and
                  Sarath Sreedharan and
                  Subbarao Kambhampati},
  title        = {Leveraging Pre-trained Large Language Models to Construct and Utilize
                  World Models for Model-based Task Planning},
  booktitle    = {NeurIPS},
  year         = {2023}
}

@inproceedings{chen-etal:iclr-2024,
  author       = {Xinyun Chen and
                  Maxwell Lin and
                  Nathanael Sch{\"{a}}rli and
                  Denny Zhou},
  title        = {Teaching Large Language Models to Self-Debug},
  booktitle    = {ICLR},
  publisher    = {OpenReview.net},
  year         = {2024}
}

@inproceedings{zhou-etal:iclr-2024,
  author       = {Aojun Zhou and
                  Ke Wang and
                  Zimu Lu and
                  Weikang Shi and
                  Sichun Luo and
                  Zipeng Qin and
                  Shaoqing Lu and
                  Anya Jia and
                  Linqi Song and
                  Mingjie Zhan and
                  Hongsheng Li},
  title        = {Solving Challenging Math Word Problems Using {GPT-4} Code Interpreter
                  with Code-based Self-Verification},
  booktitle    = {ICLR},
  publisher    = {OpenReview.net},
  year         = {2024},
}

@inproceedings{madaan-etal:nips-2023,
  author       = {Aman Madaan and
                  Niket Tandon and
                  Prakhar Gupta and
                  Skyler Hallinan and
                  Luyu Gao and
                  Sarah Wiegreffe and
                  Uri Alon and
                  Nouha Dziri and
                  Shrimai Prabhumoye and
                  Yiming Yang and
                  Shashank Gupta and
                  Bodhisattwa Prasad Majumder and
                  Katherine Hermann and
                  Sean Welleck and
                  Amir Yazdanbakhsh and
                  Peter Clark},
  title        = {Self-Refine: Iterative Refinement with Self-Feedback},
  booktitle    = {NeurIPS},
  year         = {2023}
}

@article{liu-etal:corr-2023,
  author       = {Bo Liu and
                  Yuqian Jiang and
                  Xiaohan Zhang and
                  Qiang Liu and
                  Shiqi Zhang and
                  Joydeep Biswas and
                  Peter Stone},
  title        = {{LLM+P:} Empowering Large Language Models with Optimal Planning Proficiency},
  journal      = {CoRR},
  volume       = {abs/2304.11477},
  year         = {2023},
  url          = {https://doi.org/10.48550/arXiv.2304.11477},
  eprinttype    = {arXiv},
}

@inproceedings{muennighoff-etal:iclr-2024,
  author       = {Niklas Muennighoff and
                  Qian Liu and
                  Armel Randy Zebaze and
                  Qinkai Zheng and
                  Binyuan Hui and
                  Terry Yue Zhuo and
                  Swayam Singh and
                  Xiangru Tang and
                  Leandro von Werra and
                  Shayne Longpre},
  title        = {OctoPack: Instruction Tuning Code Large Language Models},
  booktitle    = {ICLR},
  publisher    = {OpenReview.net},
  year         = {2024}
}

@inproceedings{oswald-etal:icaps-2024,
  author       = {James T. Oswald and
                  Kavitha Srinivas and
                  Harsha Kokel and
                  Junkyu Lee and
                  Michael Katz and
                  Shirin Sohrabi},
  title        = {Large Language Models as Planning Domain Generators},
  booktitle    = {ICAPS},
  pages        = {423--431},
  publisher    = {{AAAI} Press},
  year         = {2024}
}

@inproceedings{zhang-etal:acl-2023,
  author       = {Kechi Zhang and
                  Zhuo Li and
                  Jia Li and
                  Ge Li and
                  Zhi Jin},
  title        = {Self-Edit: Fault-Aware Code Editor for Code Generation},
  booktitle    = {ACL},
  pages        = {769--787},
  publisher    = {Association for Computational Linguistics},
  year         = {2023}
}

@inproceedings{zhong-etal:aclf-2024,
  author       = {Li Zhong and
                  Zilong Wang and
                  Jingbo Shang},
  title        = {Debug like a Human: {A} Large Language Model Debugger via Verifying
                  Runtime Execution Step by Step},
  booktitle    = {ACL (Findings)},
  pages        = {851--870},
  publisher    = {Association for Computational Linguistics},
  year         = {2024}
}

@article{taitler-etal:aimag-2024,
  author       = {Ayal Taitler and
                  Ron Alford and
                  Joan Espasa and
                  Gregor Behnke and
                  Daniel Fiser and
                  Michael Gimelfarb and
                  Florian Pommerening and
                  Scott Sanner and
                  Enrico Scala and
                  Dominik Schreiber and
                  Javier Segovia{-}Aguas and
                  Jendrik Seipp},
  title        = {The 2023 International Planning Competition},
  journal      = {{AI} Mag.},
  volume       = {45},
  number       = {2},
  pages        = {280--296},
  year         = {2024}
}

@article{valmeekam-etal:corr-2022,
  author       = {Karthik Valmeekam and
                  Alberto Olmo Hernandez and
                  Sarath Sreedharan and
                  Subbarao Kambhampati},
  title        = {Large Language Models Still Can't Plan {(A} Benchmark for LLMs on
                  Planning and Reasoning about Change)},
  journal      = {CoRR},
  volume       = {abs/2206.10498},
  year         = {2022},
  eprinttype    = {arXiv}
}

@article{hoffman:aimag-2001,
  author       = {J{\"{o}}rg Hoffmann},
  title        = {{FF:} The Fast-Forward Planning System},
  journal      = {{AI} Mag.},
  volume       = {22},
  number       = {3},
  pages        = {57--62},
  year         = {2001}
}

@article{aghzal2025surveylargelanguagemodels,
      title={A Survey on Large Language Models for Automated Planning}, 
      author={Mohamed Aghzal and Erion Plaku and Gregory J. Stein and Ziyu Yao},
      year={2025},
      journal      = {CoRR},
      eprinttype={arXiv},
      volume       ={abs/2502.12435}, 
}

@article{tantakoun-etal:corr-2025,
  author       = {Marcus Tantakoun and
                  Xiaodan Zhu and
                  Christian Muise},
  title        = {LLMs as Planning Modelers: {A} Survey for Leveraging Large Language
                  Models to Construct Automated Planning Models},
  journal      = {CoRR},
  volume       = {abs/2503.18971},
  year         = {2025},
}

@inproceedings{silver-etal:aaai-2024,
  author       = {Tom Silver and
                  Soham Dan and
                  Kavitha Srinivas and
                  Joshua B. Tenenbaum and
                  Leslie Pack Kaelbling and
                  Michael Katz},
  title        = {Generalized Planning in {PDDL} Domains with Pretrained Large Language
                  Models},
  booktitle    = {AAAI},
  pages        = {20256--20264},
  publisher    = {{AAAI} Press},
  year         = {2024}
}

@inproceedings{react:iclr-2023,
  author       = {Shunyu Yao and
                  Jeffrey Zhao and
                  Dian Yu and
                  Nan Du and
                  Izhak Shafran and
                  Karthik R. Narasimhan and
                  Yuan Cao},
  title        = {ReAct: Synergizing Reasoning and Acting in Language Models},
  booktitle    = {ICLR},
  publisher    = {OpenReview.net},
  year         = {2023}
}

@article{rewoo:corr-2023,
  author       = {Binfeng Xu and
                  Zhiyuan Peng and
                  Bowen Lei and
                  Subhabrata Mukherjee and
                  Yuchen Liu and
                  Dongkuan Xu},
  title        = {ReWOO: Decoupling Reasoning from Observations for Efficient Augmented
                  Language Models},
  journal      = {CoRR},
  volume       = {abs/2305.18323},
  year         = {2023}
}

@inproceedings{rap:emnkp-2023,
  author       = {Shibo Hao and
                  Yi Gu and
                  Haodi Ma and
                  Joshua Jiahua Hong and
                  Zhen Wang and
                  Daisy Zhe Wang and
                  Zhiting Hu},
  title        = {Reasoning with Language Model is Planning with World Model},
  booktitle    = {EMNLP},
  pages        = {8154--8173},
  publisher    = {Association for Computational Linguistics},
  year         = {2023}
}

@inproceedings{tot:nips-2023,
  author       = {Shunyu Yao and
                  Dian Yu and
                  Jeffrey Zhao and
                  Izhak Shafran and
                  Tom Griffiths and
                  Yuan Cao and
                  Karthik Narasimhan},
  title        = {Tree of Thoughts: Deliberate Problem Solving with Large Language Models},
  booktitle    = {NeurIPS},
  year         = {2023}
}

@inproceedings{got:aaai-2024,
  author       = {Maciej Besta and
                  Nils Blach and
                  Ales Kubicek and
                  Robert Gerstenberger and
                  Michal Podstawski and
                  Lukas Gianinazzi and
                  Joanna Gajda and
                  Tomasz Lehmann and
                  Hubert Niewiadomski and
                  Piotr Nyczyk and
                  Torsten Hoefler},
  title        = {Graph of Thoughts: Solving Elaborate Problems with Large Language
                  Models},
  booktitle    = {AAAI},
  pages        = {17682--17690},
  publisher    = {{AAAI} Press},
  year         = {2024}
}

@inproceedings{aot:icml-2024,
  author       = {Bilgehan Sel and
                  Ahmad Al{-}Tawaha and
                  Vanshaj Khattar and
                  Ruoxi Jia and
                  Ming Jin},
  title        = {Algorithm of Thoughts: Enhancing Exploration of Ideas in Large Language
                  Models},
  booktitle    = {ICML}, 
  publisher    = {OpenReview.net},
  year         = {2024}
}

@inproceedings{reflection:nips-2023,
  author       = {Noah Shinn and
                  Federico Cassano and
                  Ashwin Gopinath and
                  Karthik Narasimhan and
                  Shunyu Yao},
  title        = {Reflexion: language agents with verbal reinforcement learning},
  booktitle    = {NeurIPS},
  year         = {2023}
}

@inproceedings{tos:nips-2024,
  author       = {Michael Katz and
                  Harsha Kokel and
                  Kavitha Srinivas and
                  Shirin Sohrabi},
  title        = {Thought of Search: Planning with Language Models Through The Lens
                  of Efficiency},
  booktitle    = {NeurIPS},
  year         = {2024}
}

@inproceedings{chen-etal:aaai-2024,
  author       = {Dillon Ze Chen and
                  Sylvie Thi{\'{e}}baux and
                  Felipe W. Trevizan},
  title        = {Learning Domain-Independent Heuristics for Grounded and Lifted Planning},
  booktitle    = {AAAI},
  pages        = {20078--20086},
  publisher    = {{AAAI} Press},
  year         = {2024}
}

@inproceedings{chen-thiebaux:socs-2024,
  author       = {Dillon Z. Chen and
                  Sylvie Thi{\'{e}}baux},
  title        = {Novelty Heuristics, Multi-Queue Search, and Portfolios for Numeric
                  Planning},
  booktitle    = {SOCS},
  pages        = {203--207},
  year         = {2024}
}

@inproceedings{rintanen:aaai-20215,
  author       = {Jussi Rintanen},
  title        = {Impact of Modeling Languages on the Theory and Practice in Planning
                  Research},
  booktitle    = {AAAI},
  pages        = {4052--4056},
  publisher    = {{AAAI} Press},
  year         = {2015}
}

@inproceedings{kambhampati-etal:icml-2024,
  author       = {Subbarao Kambhampati and
                  Karthik Valmeekam and
                  Lin Guan and
                  Mudit Verma and
                  Kaya Stechly and
                  Siddhant Bhambri and
                  Lucas Saldyt and
                  Anil Murthy},
  title        = {Position: LLMs Can't Plan, But Can Help Planning in LLM-Modulo Frameworks},
  booktitle    = {ICML},
  publisher    = {OpenReview.net},
  year         = {2024}
}

@article{Anam2025PromptEA,
  title={Prompt Engineering and the Effectiveness of Large Language Models in Enhancing Human Productivity},
  author={Rizal Khoirul Anam},
  journal={ArXiv},
  year={2025},
  volume={abs/2507.18638}
}

@misc{Domains_Project,
  author = {Green and Izhaki},
  title  = {PDDL domains in rust},
  year   = {2025},
  howpublished = {\url{https://github.com/DavidIzhaki/Domains_Project/tree/11ad675ce4d5689b9a2084af36b68740b664f96d}},
  note   = {GitHub repository, commit \texttt{11ad675ce4d5689b9a2084af36b68740b664f96d}}
}




\end{document}


\maketitle

\section*{Prompt}

Below is the system prompt used in all generations and experiments, and the prompts with and without Instance Specificity. Variables in square brackets represent insertions dependent on run. [RUST\_DOMAIN\_DEFINITION] is replaced by the domain definition as encoded in rust, and [INSTANCE\_JSON] by the json.

We will note the only difference between the IS and non-IS prompts are the last few lines with the added json itself.

\subsection*{System Prompt}

\begin{lstlisting}[style=boxed, caption={}, basicstyle=\tiny, breaklines=true]
You are a senior Rust engineer specializing in heuristic design for forward-search planners.
- Always produce compiling, self-contained code.
- Never introduce types not present in the domain file.
- Prefer clarity, comments, and determinism over clever tricks.
- Heuristic values should decrease as we approach the goal (admissibility is optional).
- Helper functions must end with `_heuristic_helper` and live in the same block as the heuristic.
- When matching, include a default `_ =>` arm.
- Assume unit action costs unless the domain says otherwise.
- Output EXACTLY one Rust code block when asked for code.
\end{lstlisting}

\subsection*{Prompt}

\begin{lstlisting}[style=boxed, caption={}, basicstyle=\tiny, breaklines=true]
I have a Rust implementation of a forward-search problem. The main algorithm is complete; I need a strong heuristic.
Please invent an effective heuristic for this domain, then implement it in Rust.

Important tips/guidelines:
1) Comment your reasoning in code.
2) Be careful with numeric casts (e.g., `as f32`).
3) When using a match arm, add an emergency default arm `_ =>`.
4) If you create helper functions, end their name with `_heuristic_helper` and put them in the same code block.
5) Write the complete heuristic; do not leave placeholders. The code must be a finished product.
6) Assume action costs=1 for all actions

Write the code in the module scope (i.e. just start writing the function/helper functions, don't go into implementation scope)
The function signature must be precisely:

```rust
fn heuristic(problem: &ProblemType, state: &State) -> f64
```

(Replace `ProblemType` with the actual type from the domain.)

domain definition:
------------------
[RUST_DOMAIN_DEFINITION]
\end{lstlisting}

\subsection*{Prompt (IS)}

\begin{lstlisting}[style=boxed, caption={}, basicstyle=\tiny, breaklines=true]
I have a Rust implementation of a forward-search problem. The main algorithm is complete; I need a strong heuristic.
Please invent an effective heuristic for this domain, then implement it in Rust.

Important tips/guidelines:
1) Comment your reasoning in code.
2) Be careful with numeric casts (e.g., `as f32`).
3) When using a match arm, add an emergency default arm `_ =>`.
4) If you create helper functions, end their name with `_heuristic_helper` and put them in the same code block.
5) Write the complete heuristic; do not leave placeholders. The code must be a finished product.
6) Assume action costs=1 for all actions

Write the code in the module scope (i.e. just start writing the function/helper functions, don't go into implementation scope)
The function signature must be precisely:

```rust
fn heuristic(problem: &ProblemType, state: &State) -> f64
```

(Replace `ProblemType` with the actual type from the domain.)

domain definition:
------------------
[RUST_DOMAIN_DEFINITION]

------------------
As additional input, here's the json of the specific instance the heuristic you generate will be used to solve.
------------------
[INSTANCE_JSON]
\end{lstlisting}

\section*{Example instance and heuristic}

Below is an example instance from the Pacman domain, along with a heuristic that solved it as generated by GPT-4.1 (non-IS) and provided as-is - all comments in the code were generated by the model itself. For convinience we provide a visualization of the instance.

\subsection*{Pacman instance 16}
Below is a diagram representing the initial state. The yellow dot represents pacman, the red represents a ghost, the blue dots represent powerups, and every passable (white) space is initialized with a pellet that must be collected.
\def\PM{%
  {1,1,1,1,1,1,1,0,1,1,1,1,1,0,1,1,1,1,1,0},%
  {1,1,1,1,1,1,0,1,0,1,1,1,1,1,1,1,1,0,1,1},%
  {0,1,1,0,1,1,1,1,1,0,0,1,1,1,0,1,1,0,1,1},%
  {1,1,1,1,1,1,1,1,1,0,0,0,1,1,1,1,0,0,1,1},%
  {1,0,1,1,1,0,1,1,1,0,0,1,1,0,1,1,1,1,1,0},%
  {0,1,1,1,1,1,0,1,1,1,0,1,0,0,1,1,1,1,1,1},%
  {1,0,1,0,1,1,0,1,1,1,1,0,1,1,1,1,1,0,1,1},%
  {1,1,1,0,1,1,1,0,1,1,0,1,1,1,1,0,0,1,1,1},%
  {0,1,1,1,1,1,1,1,1,1,1,1,1,1,1,1,0,1,0,1},%
  {1,1,1,0,1,1,1,1,1,1,0,1,1,1,1,1,1,0,1,1},%
  {1,1,1,1,1,1,1,1,1,1,1,1,1,1,1,0,1,1,1,0},%
  {1,1,1,1,0,1,1,1,0,1,1,1,0,1,0,1,0,1,1,1},%
  {1,1,1,0,1,0,1,1,0,1,1,1,1,1,1,1,1,1,1,1},%
  {1,1,1,1,1,1,1,1,1,1,1,1,1,0,1,0,1,0,1,0},%
  {0,1,1,1,1,0,0,0,1,1,0,1,1,1,1,1,1,1,1,1},%
  {1,1,1,0,1,1,1,0,1,0,1,1,1,1,1,1,1,1,1,1},%
  {1,1,1,1,1,1,0,1,1,1,1,1,1,0,1,0,0,1,1,1},%
  {1,1,1,1,1,1,1,1,1,1,1,0,1,0,1,1,0,1,1,1},%
  {1,1,1,1,1,1,1,1,0,1,1,1,0,1,1,1,1,0,1,1},%
  {0,1,1,0,0,1,1,1,1,1,1,1,1,1,1,1,0,1,1,1}%
}

\def\PUs{{16/0},{2/10},{7/11},{8/17},{0/13},{17/8},{3/18},{1/15},{19/16},%
         {11/0},{4/16},{8/0},{6/15},{15/11},{11/8}}

\newcommand{\Cell}[3]{%
  \ifnum#3=1
    \fill[white] (#1,#2) rectangle ++(1,1);
  \else
    \fill[black!80] (#1,#2) rectangle ++(1,1);
  \fi
}

\begin{tikzpicture}[scale=0.25]
  \foreach \row [count=\y from 0] in \PM {
    \foreach \val [count=\x from 0] in \row {
      \Cell{\x}{19-\y}{\val}
    }
  }
  \draw[step=1cm,gray] (0,0) grid (20,20);

  \node[fill=yellow,circle,inner sep=0pt,minimum size=0.2cm]
    at ({6+0.5},{19-2+0.5}) {};

  \node[fill=red,circle,inner sep=0pt,minimum size=0.2cm]
    at ({15+0.5},{19-6+0.5}) {};

  \foreach \x/\y in \PUs {
    \node[fill=blue,circle,inner sep=0pt,minimum size=0.1cm]
      at ({\x+0.5},{19-\y+0.5}) {};
  }
\end{tikzpicture}

\begin{lstlisting}[basicstyle=\scriptsize\ttfamily,
                   breaklines=true,
                   breakatwhitespace=false,
                   columns=fullflexible]
{"passability_map":[[true,true,true,true,true,true,true,false,true,true,true,true,true,false,true,true,true,true,true,false],[true,true,true,true,true,true,false,true,false,true,true,true,true,true,true,true,true,false,true,true
],[false,true,true,false,true,true,true,true,true,false,false,true,true,true,false,true,true,false,true,true],[true,true,true,true,true,true,true,true,true,false,false,false,true,true,true,true,false,false,true,true],[true,false
,true,true,true,false,true,true,true,false,false,true,true,false,true,true,true,true,true,false],[false,true,true,true,true,true,false,true,true,true,false,true,false,false,true,true,true,true,true,true],[true,false,true,false,t
rue,true,false,true,true,true,true,false,true,true,true,true,true,false,true,true],[true,true,true,false,true,true,true,false,true,true,false,true,true,true,true,false,false,true,true,true],[false,true,true,true,true,true,true,t
rue,true,true,true,true,true,true,true,true,false,true,false,true],[true,true,true,false,true,true,true,true,true,true,false,true,true,true,true,true,true,false,true,true],[true,true,true,true,true,true,true,true,true,true,true,
true,true,true,true,false,true,true,true,false],[true,true,true,true,false,true,true,true,false,true,true,true,false,true,false,true,false,true,true,true],[true,true,true,false,true,false,true,true,false,true,true,true,true,true,true,true,true,true,true,true],[true,true,true,true,true,true,true,true,true,true,true,true,true,false,true,false,true,false,true,false],[false,true,true,true,true,false,false,false,true,true,false,true,true,true,true,true,true,true,true,true],[true,true,true,false,true,true,true,false,true,false,true,true,true,true,true,true,true,true,true,true],[true,true,true,true,true,true,false,true,true,true,true,true,true,false,true,false,false,true,true,true],[true,true,true,true,true,true,true,true,true,true,true,false,true,false,true,true,false,true,true,true],[true,true,true,true,true,true,true,true,false,true,true,true,false,true,true,true,true,false,true,true],[false,true,true,false,false,true,true,true,true,true,true,true,true,true,true,true,false,true,true,true]],"pacman_position":{"x":6,"y":2},"ghosts":{"ghost1":{"x":15,"y":6}},"power_ups":[{"x":16,"y":0},{"x":2,"y":10},{"x":7,"y":11},{"x":8,"y":17},{"x":0,"y":13},{"x":17,"y":8},{"x":3,"y":18},{"x":1,"y":15},{"x":19,"y":16},{"x":11,"y":0},{"x":4,"y":16},{"x":8,"y":0},{"x":6,"y":15},{"x":15,"y":11},{"x":11,"y":8}],"lives":5,"power_up_turns_remaining":null}
\end{lstlisting}

\subsection*{Heuristic}
\begin{lstlisting}[language=Rust, style=boxed, caption={}, basicstyle=\tiny, breaklines=true]
fn heuristic(problem: &PacmanProblem, state: &State) -> f64 {
    // --- Helper: Manhattan distance ---
    fn manhattan_distance_heuristic_helper(a: &Position, b: &Position) -> i32 {
        (a.x - b.x).abs() + (a.y - b.y).abs()
    }

    // --- Helper: Minimum Manhattan distance from Pacman to a target set ---
    fn min_distance_to_set_heuristic_helper(pos: &Position, set: &std::collections::HashSet<Position>) -> Option<i32> {
        set.iter()
            .map(|target| manhattan_distance_heuristic_helper(pos, target))
            .min()
    }
    // --- Helper: Minimum Manhattan distance from Pacman to a value set ---
    fn min_distance_to_ghosts_heuristic_helper(
        pos: &Position,
        ghosts: &std::collections::HashMap<String, Position>,
    ) -> Option<i32> {
        ghosts.values()
            .map(|g| manhattan_distance_heuristic_helper(pos, g))
            .min()
    }

    // ---- Pellet term ----
    let pellets_left = state.pellets.len() as f64;
    // Distance to nearest pellet, or 0 if all pellets already eaten (goal)
    let nearest_pellet_distance = match min_distance_to_set_heuristic_helper(&state.pacman_position, &state.pellets) {
        Some(d) => d as f64,
        None => 0.0, // No pellets left
    };

    // ---- Danger term: penalize ghost proximity if not powered up ----
    let powerup_turns = state.power_up_turns_remaining.unwrap_or(0);
    let ghost_proximity_penalty = match min_distance_to_ghosts_heuristic_helper(&state.pacman_position, &state.ghosts) {
        Some(d) => {
            if powerup_turns == 0 && d <= 2 {
                // Ghost is dangerously close and Pacman not powered up
                let penalty = (3 - d) * 50; // d=2=>50, d=1=>100, d=0=>150
                penalty as f64
            } else {
                0.0
            }
        }
        None => 0.0, // No ghosts? (Shouldn't happen)
    };

    // ---- Power-up bonus: encourage grabbing a power-up if ghost is very close ----
    let nearest_powerup_distance = match min_distance_to_set_heuristic_helper(&state.pacman_position, &state.power_ups) {
        Some(d) => d as f64,
        None => 1000.0, // No powerups left
    };
    let nearest_ghost_distance = match min_distance_to_ghosts_heuristic_helper(&state.pacman_position, &state.ghosts) {
        Some(d) => d as f64,
        None => 1000.0,
    };
    let powerup_bonus = if powerup_turns == 0 && nearest_ghost_distance <= 2.0 && nearest_powerup_distance <= 3.0 {
        -30.0
    } else {
        0.0
    };

    // ---- Sum parts together ----
    // Weighted sum: The weights can be tuned for performance, but the following is a good start:
    let pellet_term = pellets_left * 10.0 + nearest_pellet_distance;
    let heuristic_value = pellet_term + ghost_proximity_penalty + powerup_bonus;

    heuristic_value
}

\end{lstlisting}

\section{Coverage results by model and fallbacks}

Below are the results of all models with all fallback strategies.
Additional experiments and graphs can be provided to readers upon request

\begin{table*}[!t]
\centering

\setlength{\tabcolsep}{3.2pt}
\newcommand{\numtasks}[1]{\hfill (#1)}
\begin{tabular}{@{}ll|rrrrrrr|rr@{}}
\multicolumn{2}{c}{} & \multicolumn{7}{|c}{GPT} & \multicolumn{2}{|c}{Claude} \\
Domain & & 4.1 & 4.1-mini & 4.1 (IS) & 5.1 & 5-mini & 5.1=high & 5-mini=high & Sonnet 4.5 & Haiku 4.5\\
\midrule

Block Grouping & \numtasks{20} & 16 & 17 & 9 & 11 & 17 & 17 & 15  & 15 & 0 \\
Counters & \numtasks{20}       & 10 & 5 & 8 & 10 & 10 & 5 & 10  & 5 & 10 \\
Delivery & \numtasks{20}       & 16 & 13 & 10 & 16 & 20 & 16 & 20  & 16 & 0 \\
Drone & \numtasks{20}          & 15 & 16 & 15 & 16 & 16 & 16 & 16  & 16 & 15 \\
Expedition & \numtasks{20}     & 3 & 2 & 1 & 3 & 1 & 2 & 1  & 2 & 2 \\
Farming & \numtasks{20}        & 18 & 18 & 19 & 20 & 19 & 0 & 3  & 20 & 20 \\
FO-Counters & \numtasks{20}    & 4 & 4 & 8 & 5 & 6 & 6 & 6  & 6 & 4 \\
FO-Farming & \numtasks{20}     & 20 & 20 & 18 & 17 & 17 & 20 & 19  & 20 & 20 \\
FO-Sailing & \numtasks{20}     & 20 & 8 & 10 & 0 & 15 & 12 & 20  & 11 & 20 \\
Hydropower & \numtasks{20}     & 8 & 9 & 7 & 6 & 8 & 20 & 3  & 8 & 8 \\
Market Trader & \numtasks{20}  & 20 & 0 & 1 & 0 & 20 & 20 & 3  & 0 & 5 \\
Pathways & \numtasks{20}       & 0 & 1 & 1 & 0 & 0 & 1 & 1  & 0 & 1 \\
Plant Watering & \numtasks{20} & 9 & 0 & 12 & 10 & 17 & 16 & 0  & 6 & 0 \\
Rover & \numtasks{20}          & 3 & 4 & 2 & 4 & 4 & 4 & 4  & 4 & 3 \\
Sailing & \numtasks{20}        & 11 & 13 & 14 & 19 & 19 & 18 & 20  & 19 & 10 \\
Settlers & \numtasks{20}       & 1 & 1 & 2 & 1 & 1 & 0 & 1  & 3 & 1 \\
TPP & \numtasks{20}            & 13 & 0 & 11 & 0 & 4 & 4 & 20  & 0 & 0 \\
Zenotravel & \numtasks{20}     & 20 & 13 & 14 & 20 & 0 & 20 & 0  & 14 & 14 \\
\midrule
$\sum$ & \numtasks{360}        & 207 & 144 & 162 & 158 & 194 & 197 & 162  & 165 & 133 \\
\midrule
Twin Prime & \numtasks{20}     & 18 & 15 & 20 & 17 & 20 & 7 & 0  & 17 & 20 \\
Pacman & \numtasks{20}         & 9 & 6 & 13 & 13 & 9 & 14 & 16  & 10 & 15 \\
\midrule
$\sum$ & \numtasks{400}        & 234 & 165 & 195 & 188 & 223 & 218 & 178  & 192 & 168 \\

\end{tabular}
    \caption{First compilation all models.}
    \label{table-SM_FC}
\end{table*}

\begin{table*}[!t]
\centering

\setlength{\tabcolsep}{3.2pt}
\newcommand{\numtasks}[1]{\hfill (#1)}
\begin{tabular}{@{}ll|rrrrrrr|rr@{}}
\multicolumn{2}{c}{} & \multicolumn{7}{|c}{GPT} & \multicolumn{2}{|c}{Claude} \\
Domain & & 4.1 & 4.1-mini & 4.1 (IS) & 5.1 & 5-mini & 5.1=high & 5-mini=high & Sonnet 4.5 & Haiku 4.5\\
\midrule

Block Grouping & \numtasks{20} & 19 & 20 & 10 & 17 & 20 & 18 & 15 & 18 & 20 \\
Counters & \numtasks{20}       & 10 & 10 & 11 & 10 & 10 & 10 & 11 & 5 & 10 \\
Delivery & \numtasks{20}       & 20 & 19 & 18 & 16 & 20 & 17 & 20 & 20 & 20 \\
Drone & \numtasks{20}          & 16 & 16 & 16 & 16 & 16 & 16 & 16 & 17 & 16 \\
Expedition & \numtasks{20}     & 3 & 2 & 4 & 3 & 4 & 3 & 1 & 4 & 3 \\
Farming & \numtasks{20}        & 19 & 20 & 20 & 20 & 20 & 0 & 20 & 20 & 20 \\
FO-Counters & \numtasks{20}    & 4 & 6 & 14 & 6 & 6 & 6 & 6 & 6 & 6 \\
FO-Farming & \numtasks{20}     & 20 & 20 & 20 & 20 & 19 & 20 & 19 & 20 & 20 \\
FO-Sailing & \numtasks{20}     & 20 & 20 & 19 & 20 & 20 & 19 & 20 & 18 & 20 \\
Hydropower & \numtasks{20}     & 12 & 9 & 9 & 12 & 9 & 20 & 12 & 8 & 8 \\
Market Trader & \numtasks{20}  & 20 & 3 & 3 & 1 & 20 & 20 & 3 & 20 & 9 \\
Pathways & \numtasks{20}       & 1 & 1 & 1 & 1 & 1 & 1 & 1 & 1 & 1 \\
Plant Watering & \numtasks{20} & 20 & 20 & 18 & 20 & 20 & 20 & 0 & 20 & 15 \\
Rover & \numtasks{20}          & 4 & 5 & 3 & 4 & 5 & 4 & 4 & 4 & 5 \\
Sailing & \numtasks{20}        & 20 & 20 & 20 & 20 & 19 & 20 & 20 & 19 & 19 \\
Settlers & \numtasks{20}       & 1 & 2 & 3 & 1 & 2 & 0 & 1 & 3 & 2 \\
TPP & \numtasks{20}            & 16 & 0 & 13 & 4 & 4 & 20 & 20 & 2 & 0 \\
Zenotravel & \numtasks{20}     & 20 & 17 & 20 & 20 & 0 & 20 & 0 & 17 & 17 \\
\midrule
$\sum$ & \numtasks{360}        & 245 & 210 & 222 & 211 & 215 & 234 & 189 & 222 & 211 \\
\midrule
Twin Prime & \numtasks{20}     & 18 & 15 & 20 & 17 & 20 & 7 & 0 & 17 & 20 \\
Pacman & \numtasks{20}         & 9 & 16 & 13 & 14 & 16 & 16 & 16 & 16 & 16 \\
\midrule
$\sum$ & \numtasks{400}        & 272 & 241 & 255 & 242 & 251 & 257 & 205 & 255 & 247 \\

\end{tabular}
    \caption{Until success all models.}
    \label{table-SM_US}
\end{table*}

\begin{table*}
\centering

\setlength{\tabcolsep}{3.2pt}
\newcommand{\numtasks}[1]{\hfill (#1)}
\begin{tabular}{@{}ll|rrrrrrr|rr@{}}
\multicolumn{2}{c}{} & \multicolumn{7}{|c}{GPT} & \multicolumn{2}{|c}{Claude} \\
Domain & & 4.1 & 4.1-mini & 4.1 (IS) & 5.1 & 5-mini & 5.1=high & 5-mini=high & Sonnet 4.5 & Haiku 4.5\\
\midrule

Block Grouping & \numtasks{20} & 19 & 20 & 5 & 17 & 20 & 0 & 0 & 18 & 19 \\
Counters & \numtasks{20}       & 10 & 10 & 11 & 10 & 10 & 10 & 0 & 5 & 10 \\
Delivery & \numtasks{20}       & 20 & 19 & 18 & 16 & 20 & 0 & 0 & 20 & 20 \\
Drone & \numtasks{20}          & 16 & 16 & 16 & 16 & 16 & 0 & 15 & 16 & 16 \\
Expedition & \numtasks{20}     & 3 & 2 & 2 & 3 & 3 & 2 & 0 & 3 & 2 \\
Farming & \numtasks{20}        & 18 & 20 & 20 & 20 & 20 & 0 & 20 & 20 & 20 \\
FO-Counters & \numtasks{20}    & 4 & 6 & 11 & 6 & 6 & 0 & 0 & 6 & 6 \\
FO-Farming & \numtasks{20}     & 20 & 20 & 20 & 20 & 19 & 0 & 0 & 20 & 20 \\
FO-Sailing & \numtasks{20}     & 20 & 20 & 19 & 20 & 20 & 0 & 18 & 16 & 20 \\
Hydropower & \numtasks{20}     & 8 & 9 & 9 & 8 & 8 & 0 & 0 & 8 & 8 \\
Market Trader & \numtasks{20}  & 20 & 0 & 2 & 1 & 20 & 0 & 0 & 20 & 6 \\
Pathways & \numtasks{20}       & 1 & 1 & 1 & 1 & 0 & 1 & 0 & 1 & 1 \\
Plant Watering & \numtasks{20} & 20 & 20 & 18 & 19 & 20 & 0 & 0 & 20 & 14 \\
Rover & \numtasks{20}          & 4 & 4 & 2 & 4 & 5 & 0 & 0 & 4 & 5 \\
Sailing & \numtasks{20}        & 20 & 18 & 20 & 20 & 19 & 19 & 0 & 19 & 16 \\
Settlers & \numtasks{20}       & 1 & 1 & 2 & 1 & 1 & 0 & 0 & 3 & 2 \\
TPP & \numtasks{20}            & 16 & 0 & 14 & 4 & 4 & 0 & 0 & 2 & 0 \\
Zenotravel & \numtasks{20}     & 20 & 13 & 15 & 20 & 0 & 0 & 0 & 17 & 16 \\
\midrule
$\sum$ & \numtasks{360}        & 240 & 199 & 205 & 206 & 211 & 32 & 53 & 218 & 201 \\
\midrule
Twin Prime & \numtasks{20}     & 20 & 20 & 19 & 19 & 20 & 0 & 0 & 17 & 20 \\
Pacman & \numtasks{20}         & 16 & 16 & 13 & 14 & 13 & 0 & 16 & 15 & 16 \\
\midrule
$\sum$ & \numtasks{400}        & 276 & 235 & 237 & 239 & 244 & 32 & 69 & 250 & 237 \\

\end{tabular}
    \caption{SP-5 all models. At this time slicing reasoning models on high can barely finish generating any heuristic and are not usable.}
    \label{table-SM_SP5}
\end{table*}

\begin{table*}
\centering

\setlength{\tabcolsep}{3.2pt}
\newcommand{\numtasks}[1]{\hfill (#1)}
\begin{tabular}{@{}ll|rrrrrrr|rr@{}}
\multicolumn{2}{c}{} & \multicolumn{7}{|c}{GPT} & \multicolumn{2}{|c}{Claude} \\
Domain & & 4.1 & 4.1-mini & 4.1 (IS) & 5.1 & 5-mini & 5.1=high & 5-mini=high & Sonnet 4.5 & Haiku 4.5\\
\midrule

Block Grouping & \numtasks{20} & 19 & 20 & 10 & 18 & 20 & 0 & 0 & 18 & 20 \\
Counters & \numtasks{20}       & 10 & 10 & 11 & 10 & 10 & 0 & 0 & 5 & 10 \\
Delivery & \numtasks{20}       & 20 & 19 & 18 & 17 & 20 & 0 & 0 & 20 & 20 \\
Drone & \numtasks{20}          & 16 & 16 & 16 & 16 & 16 & 0 & 0 & 20 & 16 \\
Expedition & \numtasks{20}     & 3 & 2 & 4 & 3 & 4 & 0 & 0 & 4 & 3 \\
Farming & \numtasks{20}        & 19 & 20 & 20 & 20 & 19 & 0 & 0 & 20 & 20 \\
FO-Counters & \numtasks{20}    & 4 & 6 & 14 & 6 & 6 & 0 & 0 & 6 & 6 \\
FO-Farming & \numtasks{20}     & 20 & 20 & 20 & 20 & 18 & 0 & 0 & 20 & 20 \\
FO-Sailing & \numtasks{20}     & 20 & 20 & 20 & 20 & 17 & 0 & 0 & 18 & 20 \\
Hydropower & \numtasks{20}     & 12 & 9 & 9 & 20 & 12 & 0 & 0 & 8 & 8 \\
Market Trader & \numtasks{20}  & 20 & 3 & 3 & 1 & 20 & 0 & 0 & 20 & 9 \\
Pathways & \numtasks{20}       & 1 & 1 & 1 & 1 & 1 & 0 & 0 & 1 & 1 \\
Plant Watering & \numtasks{20} & 20 & 20 & 18 & 19 & 13 & 0 & 0 & 20 & 15 \\
Rover & \numtasks{20}          & 4 & 5 & 3 & 4 & 5 & 0 & 0 & 4 & 5 \\
Sailing & \numtasks{20}        & 20 & 20 & 20 & 20 & 19 & 0 & 0 & 19 & 19 \\
Settlers & \numtasks{20}       & 3 & 2 & 3 & 1 & 1 & 0 & 0 & 3 & 2 \\
TPP & \numtasks{20}            & 16 & 0 & 15 & 4 & 4 & 0 & 0 & 2 & 0 \\
Zenotravel & \numtasks{20}     & 20 & 17 & 20 & 20 & 0 & 0 & 0 & 17 & 16 \\
\midrule
$\sum$ & \numtasks{360}        & 247 & 210 & 225 & 220 & 205 & 0 & 0 & 225 & 210 \\
\midrule
Twin Prime & \numtasks{20}     & 18 & 20 & 20 & 20 & 20 & 0 & 0 & 17 & 20 \\
Pacman & \numtasks{20}         & 16 & 16 & 14 & 16 & 16 & 0 & 0 & 16 & 16 \\
\midrule
$\sum$ & \numtasks{400}        & 281 & 246 & 259 & 256 & 241 & 0 & 0 & 258 & 246 \\

\end{tabular}
    \caption{SP-10 all models. At this time slicing reasoning models on high can barely finish generating any heuristic and are not usable. This is the optimal configuration for almost all non-high-reasoning models.}
    \label{table-SM_SP10}
\end{table*}

\FloatBarrier
\newpage

\section{SP-N Hyperparameter Selection}

\textsc{SelfPortfolio-N (SP-N)} split the time budget equally across $N$ slices, each generating and running a different heuristic, allowing diverse heuristics at the cost of per-slice time budget. We selected N=10 as a round number to avoid post-optimization, giving each slice a total time of 1 minute. Here we analyze the effect of this choice.

Each slice is completely independent, and any additional time budget left for a slice is discarded once the search is over, be that by \textsc{success}, \textsc{OutOfMemory} (OOM), or \textsc{CompilationError}. Assuming we ignore API costs and optimize only for success rate, the best strategy to select $N$ is to take the largest $N$ such that the time budget does not significantly harm slice success rate.

Formally, define $N$ is the number of slices we portfolio over, and $T$ to be our full time budget. Let $X$ be the random variable taking the value of the total runtime (generation + search) for successful slices, or $T$ for unsuccessful ones, and assume that it is well approximated by $X_{our}$, the distribution of $X$ marginalized over our domains. The number of slices maximizing the coverage over the data therefore can be approximated by 

\[N^*\coloneqq\arg\!\max_{N\in\mathbb{N}} {1 - (1 - P[X_{our} < T/N])^N},\]

that is, the $N$ maximizing the probability that at least one slice is successful. $N^*$ is finite and well-defined because generation requires at least some $\varepsilon>0$ time so for $N>T/\varepsilon$ the success probability drops to zero. To approximate the random variable $X_{our}$, we decompose it into $\{X_{domain_i}\}$, its marginalizations per domain. For each domain $i$ we approximate $X_{domain_i}$ by sampling 40 heuristics and running them to completion with the full time budget. We later perform a Monte-Carlo simulation of \textsc{SP-N} over possible samplings of these heuristic functions.

Since success rate, generation time, and run time can very greatly per domain, $N^*$ may also be different per domain. To account for this and for better generalization to new domains, we present two methods to select $N$. Both methods use Monte-Carlo simulations of \textsc{SP-N} with GPT-4.1.

The first method is \textit{sum coverage intervalling}: We plot the median, and first- and third-quartiles of the coverage in Monte-Carlo simulation of \textsc{SP-N} across all domains. This most directly answers the question \textit{'which N value would achieve the best result on the benchmark'}. As seen below, this method would select $N^*\in\{11, 12\}$ but finds all values $8...13$ to be roughly equal.

The second method is \textit{common median}: We find, per domain, all values of $N$ for which the median simulation's coverage is optimal and co-equal, then plot how common each value of N is across domains. This provides a clearer domain breakdown and would allow us for example to detect if $N^*$ is bimodal across domains. As seen below, according to this method $N^*\in\{10, 13\}$, but as before a large range of values are measured to be roughly equally effective. Notice especially that despite the domains in the IPC varying greatly in difficulty, a vast majority of the domains find similar values to be optimal.

Overall, the analysis verifies that \textsc{sp-N} is robust to a wide range of $N$ values.

\begin{figure}[ht]
    \centering
    \includegraphics[width=0.47\textwidth]{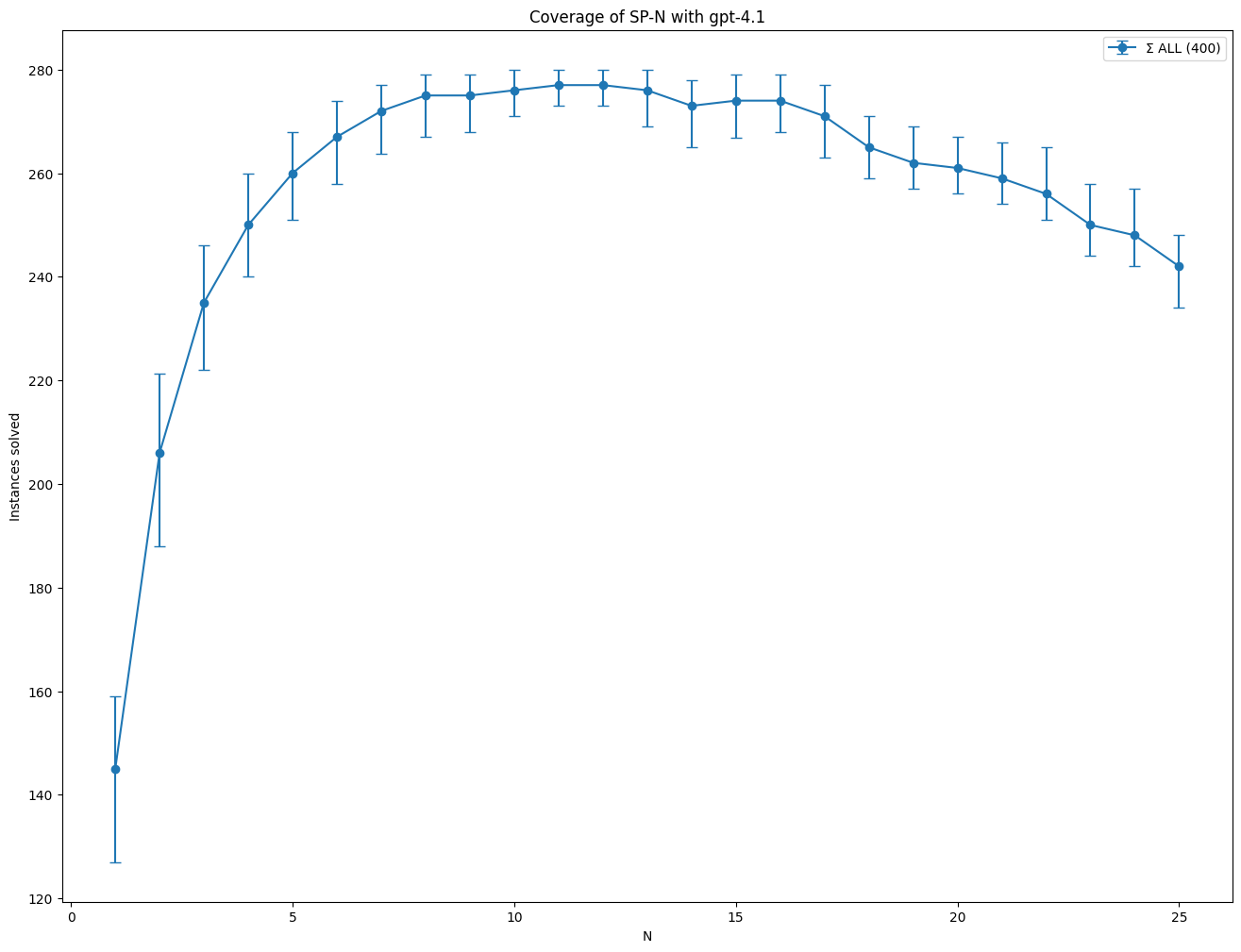}
    \hfill
    \includegraphics[width=0.47\textwidth]{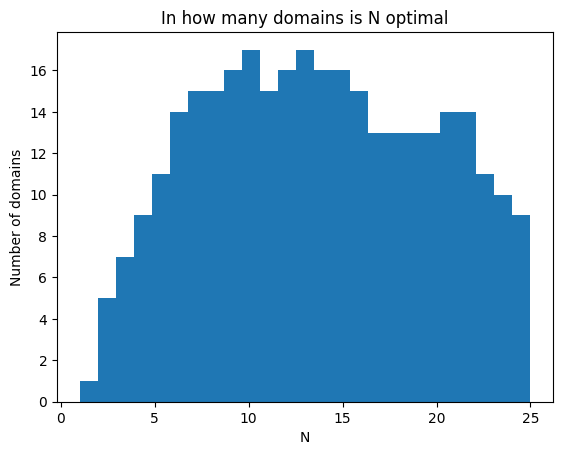}
    \caption{The coverage median coverage, with first- and third-quartile confidence bounds, of \textsc{sp-N} with GPT-4.1 according to Monte-Carlo simulations for possible values of N (top). In how many of the 20 domains does a given value of N yield an optimal median simulation, based on Monte-Carlo simulations of \textsc{sp-N} with GPT-4.1. (bottom)}
    \label{fig:sp_n_SCI}
\end{figure}


